\documentclass[a4paper]{article}

\usepackage{graphicx}
\usepackage{authblk}
\usepackage{amsmath}
\usepackage{amsthm}
\usepackage{amssymb}
\usepackage{bm}
\usepackage{booktabs}
\usepackage{xspace}
\usepackage[margin=1.25in]{geometry}
\usepackage[numbers]{natbib}
\usepackage{hyperref}
\usepackage{tikz}
\usepackage{dsfont}
\usetikzlibrary{positioning, shapes.geometric}

\newcommand{\lf}{\!\!:}
\newcommand{\rf}{|}
\newcommand{\bs}{\mathbf{s}}
\newcommand{\bq}{\mathbf{q}}
\newcommand{\br}{\mathbf{r}}
\newcommand{\bx}{\mathbf{x}}
\newcommand{\be}{\mathbf{e}}
\newcommand{\bp}{\mathbf{p}}
\newcommand{\cE}{\mathcal{E}}
\newcommand{\cK}{\mathcal{K}}
\newcommand{\balpha}{\bm{\alpha}}
\newcommand{\bet}{\bm{\eta}}
\newcommand{\bdelta}{\bm{\delta}}

\newcommand{\supp}{\mathrm{supp}}
\newcommand{\free}{\mathrm{free}}
\newcommand{\bzero}{\mathbf{0}}
\newcommand{\bone}{\mathds{1}}

\newcommand{\statei}[1]{\bx^{(#1)}}

\numberwithin{equation}{section}

\title{State Algebra for Probabilistic Logic}
\author[1]{Dmitry Lesnik}
\author[1,2]{Tobias Schäfer}
\affil[1]{Stratyfy Inc., New York, NY, USA}
\affil[2]{Department of Mathematics, College of Staten Island, Staten Island, NY, USA \& Physics Program, CUNY Graduate Center, NY, USA}

\date{}

\begin{document}

\maketitle

\begin{abstract}
    This paper presents a \textit{Probabilistic State Algebra} as an extension of deterministic propositional logic, providing a computational framework for constructing \textit{Markov Random Fields} (MRFs) through pure linear algebra. By mapping logical states to real-valued coordinates interpreted as energy potentials, we define an energy-based model where global probability distributions emerge from coordinate-wise Hadamard products. This approach bypasses the traditional reliance on graph-traversal algorithms and compiled circuits, utilising $t$-objects and wildcards to embed logical reduction natively within matrix operations.

    We demonstrate that this algebra constructs formal Gibbs distributions, offering a rigorous mathematical link between symbolic constraints and statistical inference. A central application of this framework is the development of \textit{Probabilistic Rule Models} (PRMs), which are uniquely capable of incorporating both probabilistic associations and deterministic logical constraints simultaneously. These models are designed to be inherently interpretable, supporting a human-in-the-loop approach to decisioning in high-stakes environments such as healthcare and finance. By representing decision logic as a modular summation of rules within a vector space, the framework ensures that complex probabilistic systems remain auditable and maintainable without compromising the rigour of the underlying configuration space.
\end{abstract}


\section{Introduction}\label{sec:introduction}

The integration of symbolic logic with probabilistic reasoning is a central theme in modern artificial intelligence. Foundational frameworks such as Markov Logic Networks (MLNs), pioneered by Richardson and Domingos~\cite{richardson2006markov}, unified first-order logic with graphical models by treating logical formulas as templates for Markov networks. In parallel, the development of Probabilistic Graphical Models (PGMs) by Koller and Friedman~\cite{koller2009probabilistic} established a rigorous methodology for handling uncertainty through Bayesian and Markov networks. While these approaches have been transformative, their practical implementations have historically relied on knowledge compilation into secondary graph-based structures, such as Reduced Ordered Binary Decision Diagrams (ROBDDs)~\cite{Bryant1986,Bryant1992} or Sentential Decision Diagrams (SDDs)~\cite{darwiche2011,choi2019}. These established paradigms depend fundamentally on graph traversals and node caching, which often become computational bottlenecks for exact inference in high-dimensional spaces.

A key computational task in these frameworks is Weighted Model Counting (WMC), which generalises the problem of counting satisfying assignments by assigning weights to literals or models~\cite{sang2005, chavira2008probabilistic}. WMC has emerged as a powerful unifying abstraction for probabilistic inference, yet its performance is often tethered to the efficiency of the underlying circuit representation. While Algebraic Model Counting (AMC) has generalised WMC to arbitrary semirings to unify various inference tasks~\cite{KIMMIG201746}, its standard implementations still rely on the underlying structural traversals of compiled decision diagrams.

Building upon the deterministic State Algebra for Propositional Logic introduced by Lesnik and Schäfer~\cite{lesnik2025}, this paper presents a Probabilistic State Algebra. This framework shifts the computational paradigm from graph theory to pure linear algebra. By mathematically representing state vectors as sums of t-objects and replacing complex graph transformations with the pairwise multiplication of ultra-sparse row vectors, the framework introduces a highly parallelisable approach to WMC\@.

The algebra adopts an energy-based interpretation where real-valued state coordinates function as energy potentials, and the exponential mapping acts as a Boltzmann factor to generate a formal Gibbs distribution. Because this framework natively functions as an Energy-Based Model (EBM) over discrete logical spaces, we can leverage these formal Gibbs distributions and zero-temperature limits to seamlessly transition between probabilistic inference and deterministic logic. Furthermore, by relying on matrix-based operations, we offer a scalable roadmap for probabilistic logic that bypasses the rigid variable-ordering constraints typically associated with canonical decision diagrams.

Recent advancements map WMC to tensor network contractions to leverage parallel computing architectures~\cite{dudek2021}. However, their computational efficiency depends heavily on finding optimal contraction orderings via graph decomposition. In contrast, Probabilistic State Algebra operates directly on the state space. By utilising wildcards and t-objects, it natively embeds logical reduction within pairwise algebraic operations, bypassing these structural graph dependencies entirely.

Beyond pure algorithmic efficiency, the shift towards a rule-based algebraic structure addresses a growing need for transparency and interpretability in automated decision-making. Rule-based models are inherently modular and permit a low-dimensional representation of complex constraints, making the underlying logic accessible to human audit. By formalising these rules within a probabilistic algebra, we provide a mathematical foundation that maintains the transparency of symbolic logic while capturing the nuance of statistical uncertainty. This combination is vital in high-stakes environments, where verifying model behaviour is as critical as predictive accuracy.

For instance, such transparency is essential when predicting the onset of viral symptoms in healthcare~\cite{dhaese-et-al:2021} or constructing diagnostic layers in finance~\cite{lesnik2025probabilistic}. In these practical cases, the models often remain low-dimensional, allowing for the seamless incorporation of expert knowledge alongside data-driven insights. Furthermore, industry applications currently relying on traditional Business Rule Engines (BREs) can transition to this probabilistic approach to significantly improve robustness. Within our framework, the arduous problem of manually checking the logical consistency of a deterministic BRE is eliminated, as the algebra natively handles conflicting or overlapping constraints through its probabilistic weighting. Consequently, the development and maintenance of \textit{Probabilistic Rule Models} (PRMs) become substantially simpler, offering a rigorous and flexible solution for modern enterprise decisioning.

The remainder of this paper is structured as follows. Section~\ref{sec:basic-definitions} briefly reviews the foundational definitions of the state space and coordinate representation. Section~\ref{sec:probabilistic-model} introduces the core probabilistic models, demonstrating how state vectors generate formal Gibbs distributions. Section~\ref{sec:probabilistic-logic} formalises the probabilistic logic framework and the computation of partition functions for systems of weighted rules. Section~\ref{sec:optimisation} discusses computational strategies and optimisations, including the use of Markov Blankets and vertex separators. Section~\ref{sec:probabilistic-rule-models} presents Probabilistic Rule Models (PRMs) as a practical application for transparent decision-making. Finally, Section~\ref{sec:conclusion} concludes with a summary and an outlook on future research directions.

\section{Basic Definitions}\label{sec:basic-definitions}

This section briefly reviews the foundational definitions introduced in~\cite{lesnik2025}. This establishes a self-contained framework.

\paragraph{State Space.} We consider a finite set of Boolean variables $\cE = \{X_1,\ldots,X_N\},\; N < \infty$. Each truth assignment
\begin{equation*}
    \bx = \{ x_1,\ldots,x_N\}\,,\qquad x_i\in\{1,0\}
\end{equation*}
is called a \textbf{state}. The \textbf{state space} $S_\cE$, also referred to as the \textbf{configuration space}, denotes the set of all possible states. It possesses $N$ degrees of freedom and contains $M=2^N$ elements.

A \textbf{state vector} is an arbitrary subset of $S_\cE$. In matrix notation, we denote a state vector as
\begin{equation*}
    \bs =
    \begin{Bmatrix}
        \statei{1}_1 & \statei{1}_2 & \ldots & \statei{1}_N \\
        \statei{2}_1 & \statei{2}_2 & \ldots & \statei{2}_N  \\
        \cdots & & & \\
        \statei{m}_1 & \statei{m}_2 & \ldots & \statei{m}_N
    \end{Bmatrix} \,,\qquad 0 \leq m \leq M
\end{equation*}
where $\statei{i}$ is the $i$-th truth assignment. The state vectors in the \textbf{set representation} form a space of all possible subsets of $S_\cE$. We designate this space as
\begin{equation*}
    \mathbb{S}^*_\cE = \mathcal{P}(S_\cE)
\end{equation*}
where $\mathcal{P}()$ is the power set. We will be using bold Latin letters $\bp,\bq,\br,\bs$ to designate state vectors.

\paragraph{Compact Notation.}
We employ ``\emph{wildcards},'' denoted by dashes ``--'', to compactly represent state vectors in matrix notation. A row containing wildcards represents the set of states generated by all possible binary substitutions (0s and 1s) in the wildcard positions. For instance:
\begin{equation*}
    \begin{Bmatrix}
        1 & - & -  \\
        0 & 0 & -
    \end{Bmatrix} =
    \begin{Bmatrix}
        1 & 0 & 0  \\
        1 & 0 & 1  \\
        1 & 1 & 0  \\
        1 & 1 & 1  \\
        0 & 0 & 0  \\
        0 & 0 & 1
    \end{Bmatrix}.
\end{equation*}
The transformation introducing wildcards is called \emph{reduction}, while the inverse process is \emph{expansion}. Generally, the reduced form of a state vector is not unique; the result of reduction may depend on the specific order in which wildcards are introduced.

A variable $X_i$ is called \textbf{free} in the state vector $\bs$ if a compact matrix representation of $\bs$ exists in which the $i$-th column contains only wildcards. Variables that are not free are called \textbf{causal}. A set of causal variables of $\bs$ is called \textbf{support}, denoted as
\begin{equation*}
    \supp(\bs)
\end{equation*}
We will denote the set of free variables as
\begin{equation*}
    \free(\bs)
\end{equation*}
Two state vectors are called \textbf{independent} if they have non-overlapping support. We designate this as follows:
\begin{align*}
    & \bs \wr \bq \quad \iff \quad \supp(\bs)\cap\supp(\bq) = \varnothing
\end{align*}

In compact matrix notation, state vectors are equivalent to a DNF representation of logical functions: a function is represented as a disjunction of clauses, whereas each clause is a conjunction of literals. In this mapping, a row of a matrix is equivalent to a product term. A single row in this compact matrix representation is formally referred to as a \textbf{\textit{t}-object}~\cite{lesnik2025}. For instance, a $t$-object (row) $\{1\;-\;0\}$ is equivalent to $X_1 \wedge \neg X_3$. The entire state vector represents a set of states satisfying the function. Wildcards denote a localised unconstrained entry within a specific row. In logic synthesis, this is analogous to a ``don't care'' condition for a specific product term. Note that a wildcard applies only to its specific row. If a variable's column consists entirely of wildcards, the overall logical function is invariant with respect to it, and we define it as a free variable.

\paragraph{Coordinate Representation.}
In compact representation, it is easy to overlook state duplicates. For instance, in the state vector
\begin{equation*}
    \begin{Bmatrix}
        1 & -   \\
        - & 0
    \end{Bmatrix}
\end{equation*}
the state $\{1\ 0\}$ is present in both rows. In a standard set representation, state duplicates are ignored.

Suppose, instead, we wish to track the multiplicity of each state within the vector. Let us systematically enumerate all possible states using the notation:
\begin{equation*}
    \statei{i}\,,\quad i = 1,\ldots,M
\end{equation*}
to designate the $i$-th state. Let $p^i$ denote the counter for the $i$-th state within the vector~$\bp$. We can equivalently express a state vector in terms of these state counters $(p^i)_{i=1}^M$.

Consider an example. Let's enumerate the states of a state space for a system of two variables as follows:
\begin{equation*}
    \begin{array}{ccc}
        \statei{1}: & 0 & 0\\
        \statei{2}: & 0 & 1\\
        \statei{3}: & 1 & 0\\
        \statei{4}: & 1 & 1
    \end{array}
\end{equation*}
The following examples illustrate the equivalent representation of state vectors as vectors of state counters:
\begin{equation*}
    \begin{Bmatrix}
        0 & 0 \\
        1 & 1
    \end{Bmatrix} = (1,\, 0,\, 0,\, 1)\,, \qquad
    \begin{Bmatrix}
        0 & 1 \\
        1 & -
    \end{Bmatrix} = (0,\, 1,\, 1,\, 1)
\end{equation*}

Now we generalise this definition by allowing values $p^i$ to be arbitrary \emph{real numbers}. A representation of a state vector in terms of real-valued vector $(p^i)$ is called \textbf{coordinate representation}, with $p^i$ playing the role of the \emph{coordinates}:
\begin{equation*}
    \bp = (p^1,p^2,\ldots,p^M) = (p^i)_{i=1}^M\,,\qquad p^i \in \mathbb{R}
\end{equation*}

Notice that this definition is more generic than the coordinate representation introduced in~\cite{lesnik2025}, where coordinates were integer-valued. We designate
\begin{equation*}
    \mathbb{S}_\cE
\end{equation*}
the space of all state vectors in coordinate representation.

We will be using \emph{upper indices} for coordinate-wise elements. For instance, for the coordinate representation of a state vector we will use the notation $\bs = (s^1,s^2,\ldots,s^M) = (s^i)_{i=1}^M$. We will use $C^i(\bs)$ to designate the $i$-th coordinate of the vector $\bs$:
\begin{align*}
    & \bs = (s^1,s^2,\ldots,s^M)\,,\qquad C^i(\bs) = s^i
\end{align*}

An empty vector $\bzero$ has all its coordinates equal to 0:
\begin{align*}
    & \bzero = (0,\,0, \ldots,\,0)\,,\qquad C^i(\bzero) = 0\quad \forall i
\end{align*}
We will use $\bone$ to designate the state vector with all coordinates equal to 1:
\begin{align*}
    & \bone = (1,\,1, \ldots,\,1)\,,\qquad C^i(\bone) = 1\quad \forall i
\end{align*}
Notice that the vector $\bone$ can be represented in matrix notation as a single row which contains only wildcards:
\begin{align*}
    \bone = \begin{Bmatrix}
                -& -&\ldots & -
    \end{Bmatrix}
\end{align*}

States having zero coordinates are called \textbf{ground states}. The vector $\bzero$ contains ground states only.

\paragraph{Algebra of State Vectors.}
We introduce coordinate-wise operations of addition, subtraction, multiplication and scaling. For two state vectors
\begin{align*}
    & \bs = (s^1, s^2,\ldots,s^M)\,,\quad C^i(\bs) = s^i\,,\quad s^i\in \mathbb{R}\,,\\
    & \bq= (q^1, q^2,\ldots,q^M)\,,    \quad C^i(\bq) = q^i\,,\quad q^i\in \mathbb{R}
\end{align*}
the algebraic operations are defined as follows:
\begin{align*}
    &\bs \, \bq= (s^1\, q^1, s^2\,q^2,\ldots, s^M\,q^M)\,, \\
    & \bs + \bq = (s^1 + q^1, s^2 + q^2,\ldots, s^M + q^M)\,, \\
    & \bs - \bq = (s^1 - q^1, s^2 - q^2,\ldots, s^M - q^M)\,, \\
    & k\,\bs = (k\,s^1, k\,s^2,\ldots, k\,s^M)\,,\qquad k\in \mathbb{R}
\end{align*}

Note some obvious identities:
\begin{align*}
    & \bs \cdot \bone = \bs \\
    & \bs \cdot \bzero = \bzero \\
    & \bs + \bzero = \bs
\end{align*}

The coordinate-wise operations turn $\mathbb{S}_\cE$ into a commutative algebra over the field of real numbers equipped with the Hadamard (coordinate-wise) product.

\paragraph{Hybrid Notation.} We occasionally employ a hybrid notation that uses matrix notation for state vectors in coordinate representation:
\begin{equation*}
    \begin{Bmatrix}
        a \; \lf & 0 & 0 & -   \\
        b \; \lf & 1 & - & 0
    \end{Bmatrix}
\end{equation*}
In this example, $a$ is the coordinate value assigned to every state contained in the first row, and $b$ is assigned to those in the second. States absent from the hybrid notation implicitly have zero coordinates. States present multiple times have their coordinates summed across all instances.

Consider an example. Let
\begin{align*}
    \bs = \begin{Bmatrix}
              1 \; \lf & 0 & - \\
              1 \; \lf & - & 0
    \end{Bmatrix}, \quad
    \bq = \begin{Bmatrix}
              1 \; \lf & 0 &  0
    \end{Bmatrix}
\end{align*}
Then in hybrid notation, for $a, b\in\mathbb{R}$:
\begin{align*}
    & a\,\bs  = \begin{Bmatrix}
                    a \; \lf & 0 & -   \\
                    a \; \lf & - & 0
    \end{Bmatrix} =
    \begin{Bmatrix}
        2\,a & \lf & 0 & 0   \\
        a    & \lf & 0 & 1   \\
        a    & \lf & 1 & 0   \\
        0    & \lf & 1 & 1
    \end{Bmatrix}, \\
    & b\,\bq  = \begin{Bmatrix}
                    b \; \lf & 0 & 0
    \end{Bmatrix} =
    \begin{Bmatrix}
        b \; \lf & 0 & 0   \\
        0 \; \lf & 0 & 1   \\
        0 \; \lf & 1 & 0   \\
        0 \; \lf & 1 & 1
    \end{Bmatrix}, \\
    & a\,\bs + b\,\bq  =
    \begin{Bmatrix}
        2\,a+b & \lf & 0 & 0   \\
        a      & \lf & 0 & 1   \\
        a      & \lf & 1 & 0   \\
        0      & \lf & 1 & 1
    \end{Bmatrix}, \\
    & \bs\,\bq = \begin{Bmatrix}
                     2 \; \lf &  0 & 0
    \end{Bmatrix}
\end{align*}

\paragraph{Binary State Vectors.}
A state vector in coordinate representation is called \textbf{binary} if all its coordinates are strictly zeros or ones.
\begin{equation*}
    C^i(\bs)\in\{0, 1\}
\end{equation*}
For instance
\begin{equation*}
    \bs = \begin{Bmatrix}
              1 \; \lf & 1 & -    \\
              1 \; \lf & 0 & 1
    \end{Bmatrix} =
    \begin{Bmatrix}
        1 \; \lf & 1 & 1    \\
        1 \; \lf & 1 & 0    \\
        1 \; \lf & 0 & 1    \\
        0 \; \lf & 0 & 0
    \end{Bmatrix}
\end{equation*}

Binary state vectors possess a one-to-one mapping with the standard set representation. Consequently, we often adopt ``set'' terminology when discussing them. In particular, we say that a state is present in the state vector if its coordinate is 1, and missing if its coordinate is 0:
\begin{align*}
    & \statei{i} \in\bs\quad \iff \quad s^i = 1 \\
    & \statei{i} \not \in\bs\quad \iff \quad s^i = 0
\end{align*}

The product of two binary vectors is also binary. It corresponds to the intersection of the two state vectors in the set representation. When computing this product using compact matrix notation, the operation is evaluated via the pairwise intersection of their respective rows ($t$-objects). The column-wise rules for intersecting individual entries are straightforward: a wildcard intersecting with a numeric value ($0$ or $1$) yields that numeric value; two wildcards yield a wildcard; and conflicting numeric values ($0$ and $1$) yield an empty intersection (meaning the rows are orthogonal and their product is $\bzero$). For a simple example using single $t$-objects:
\begin{align*}
    & \bs = \begin{Bmatrix}
                1 \; \lf & 1 & - & -
    \end{Bmatrix}\,, \quad
    \bq = \begin{Bmatrix}
              1 \; \lf & - & - & 0
    \end{Bmatrix} \\
    & \bs\,\bq = \begin{Bmatrix}
                     1 \; \lf & 1 & - & 0
    \end{Bmatrix}
\end{align*}
More generally, for state vectors containing multiple rows, the product is obtained by intersecting all possible pairs of rows between the two vectors:
\begin{align*}
    & \bs = \begin{Bmatrix}
                1 \; \lf & 1 & -    \\
                1 \; \lf & 0 & 1
    \end{Bmatrix} = \begin{Bmatrix}
                        1 \; \lf & 1 & 1    \\
                        1 \; \lf & 1 & 0    \\
                        1 \; \lf & 0 & 1    \\
                        0 \; \lf & 0 & 0
    \end{Bmatrix}\\
    & \bq = \begin{Bmatrix}
                1 \; \lf & - & 0    \\
                1 \; \lf & 0 & 1
    \end{Bmatrix}=\begin{Bmatrix}
                      0 \; \lf & 1 & 1    \\
                      1 \; \lf & 1 & 0    \\
                      1 \; \lf & 0 & 1    \\
                      1 \; \lf & 0 & 0
    \end{Bmatrix} \\
    & \bs\,\bq = \begin{Bmatrix}
                     0 \; \lf & 1 & 1    \\
                     1 \; \lf & 1 & 0    \\
                     1 \; \lf & 0 & 1    \\
                     0 \; \lf & 0 & 0
    \end{Bmatrix} = \begin{Bmatrix}
                        1 \; \lf & 1 & 0    \\
                        1 \; \lf & 0 & 1
    \end{Bmatrix}
\end{align*}
The square of a binary state vector equals the state vector itself:
\begin{align*}
    & \bs\,\bs = \bs
\end{align*}
Binary state vectors are called \textbf{orthogonal} if their product is $\bzero$. In terms of sets of states, orthogonal state vectors represent disjoint sets.

For binary state vectors we can introduce operations equivalent to set operations of union $\cup$, intersection $\cap$ and subtraction $\smallsetminus$ as follows:
\begin{align}
    & \bs \cap \bq = \bs\,\bq                   \\
    & \bs \cup \bq = \bs + \bq - \bs\,\bq       \\
    & \bs \smallsetminus \bq = \bs - \bs \,\bq
\end{align}
For the binary operands, the results of these operations are also binary.

Obviously, for any binary $\bs$:
\begin{align*}
    & \bs \cap \bzero = \bzero              \\
    & \bs \cap \bone = \bs                  \\
    & \bs \cup \bzero = \bs                 \\
    & \bs \cup \bone = \bone                \\
    & \bs \smallsetminus \bzero = \bs       \\
    & \bs \smallsetminus \bone = \bzero
\end{align*}
Notice also that the sum of binary state vectors is not binary unless they are mutually orthogonal.

The number of states present in a binary state vector is referred to as its \textbf{cardinality}, denoted as
$|\bs|$. We can formally define cardinality using the standard scalar product:
\begin{align*}
    & \langle \bs, \bq \rangle = \sum_{i=1}^M s^i\,q^i
\end{align*}
The cardinality of a state vector $\bs$ follows as
\begin{align*}
    & |\bs| = \langle \bs, \bs \rangle
\end{align*}
Making use of the fact that for binary vectors $(s^i)^2 = s^i$, we can also find the cardinality of intersection of two state vectors as
\begin{align*}
    |\bs\,\bq| = \langle \bs, \bq \rangle
\end{align*}

In what follows, we assume that state vectors designated by bold letters ($\bq,\br,\bs$) are binary unless stated otherwise.

\paragraph{Exponential Factors.}
Alongside state coordinates, we frequently utilise \textbf{exponential factors}. For the vector $\bp = (p^1, \ldots, p^N)$, the exponential factors $\psi^i$ are defined as
\begin{align}
    \psi^i = \exp(-p^i)\,,\quad \psi^i > 0
\end{align}
For exponential factors, we will use the following hybrid notation. For a binary vector $\bs$:
\begin{equation*}
    a\,\bs = \bs | \psi =
    \begin{Bmatrix}
        a \; \lf & 1 & - & 0   \\
        a \; \lf & 1 & 0 & -
    \end{Bmatrix} =
    \begin{Bmatrix}
        1 & - & 0 & \rf \; \psi  \\
        1 & 0 & - & \rf \; \psi
    \end{Bmatrix},\qquad \text{where} \quad \psi = \exp(-a)
\end{equation*}
Addition of coordinates corresponds to multiplication of exponential factors:
\begin{align}
    \label{eq:multiplication_psi_factors}
    & a\,\bs + b\,\bs = (a+b)\,\bs = \bs|\phi + \bs|\psi = \bs | \phi\,\psi \\
    & \text{where}  \nonumber \\
    & \phi = \exp(-a),\quad \psi = \exp(-b), \quad \phi\,\psi = \exp(-(a+b)) \nonumber
\end{align}

Notice that ground states (specifically, states not explicitly present in the matrix notation) inherently possess zero coordinates, corresponding to unit exponential factors. Among states with non-negative coordinates, the ground states attain the maximum exponential factor of 1.

We will be using interchangeably the following notation
\begin{align*}
    &\bp = \bs|\phi + \bq|\psi =
    \begin{Bmatrix}
        \bs & \rf\; \phi \\
        \bq & \rf\; \psi
    \end{Bmatrix}
\end{align*}
where $\bs$ and $\bq$ are binary.

\section{Probabilistic Models}\label{sec:probabilistic-model}

Having established $\mathbb{S}_{\mathcal{E}}$ as a continuous algebraic space, we now demonstrate how state vectors can natively represent probabilistic models over a logical configuration space. To do so, we apply an energy-based interpretation analogous to statistical mechanics.

In this probabilistic context, we interpret the real-valued coordinate $p^i$ of the $i$-th state $\statei{i}$ as its state energy (or energy potential). Consequently, the exponential mapping introduced in Section~\ref{sec:basic-definitions}, $\psi^i = \exp(-p^i)$, functions mathematically as the Boltzmann factor (evaluated at a normalised unit temperature). These Boltzmann factors represent the unnormalised probability mass of each state in the logical system.

By normalising these factors by the sum over all states, $Z = \sum \psi^j$, which we identify as the partition function, the state vector $\bp$ generates a formal Gibbs distribution over the state space~$S_{\mathcal{E}}$.

\subsection{Probability Distributions}
\label{sec:probability-distribution}

Consider a set of $N$ Boolean variables $\cE=\{X_1, \ldots, X_N\}$. Let $\bp$ be a state vector in coordinate representation, which has the exponential factor $\psi^i$ assigned to the $i$-th state:
\begin{align}
    \label{eq:prob_mass_coordinatewise}
    \bp = \begin{Bmatrix}
              \statei{1} & \rf & \psi^1  \\
              \statei{2} & \rf & \psi^2  \\
              \vdots \\
              \statei{M} & \rf & \psi^M
    \end{Bmatrix}
\end{align}
where $M = 2^N$ and the states $\statei{i}\in S_\cE,\ i = 1,\ldots,M$, represent all possible truth assignments. We say that a state vector $\bp$ defines a \textbf{probability distribution} over the configuration space~$S_\cE$ of the Boolean variables $\cE$. The probability of the state $\statei{i}$ is defined as
\begin{align}
    & P_\bp(\statei{i}) = \frac{\psi^i}{Z}\,,\qquad \text{where} \quad Z = \sum_{j=1}^M \psi^j
\end{align}
Here $P_\bp$ stands for probability under the distribution $\bp$. The exponential factors $\psi^i$ play a role of unnormalised probabilities, and $Z$ is the normalisation constant, also referred to as \emph{partition function}.

The same probability expressed in terms of the coordinates $p^i=-\ln(\psi^i)$ becomes
\begin{align*}
    & P_\bp(\statei{i}) = \frac{\exp(-p^i)}{Z}\,,\qquad \text{where} \quad Z = \sum_{j=1}^M \exp(-p^j)
\end{align*}

For \emph{unnormalised} probability we will be using notation
\begin{align*}
    & \hat P_\bp(\statei{i}) = \psi^i
\end{align*}

Probability of a \textbf{binary} state vector is defined as the sum of probabilities of the states it comprises. Let
\begin{align}
    & \bs =
    \begin{Bmatrix}
        \statei{i_1} \\
        \statei{i_2} \\
        \vdots \\
        \statei{i_k}
    \end{Bmatrix}
\end{align}
Then the unnormalised probability of $\bs$ is given by the sum
\begin{align}
    \hat P_\bp(\bs) = \hat P_\bp(\statei{i_1}) + \ldots + \hat P_\bp(\statei{i_k}) = \psi^{i_1} + \ldots + \psi^{i_k}
\end{align}

The unit state vector $\bone = \{-\;-\;\ldots\;- \}$ contains all possible states, and hence the partition function can be expressed as
\begin{align*}
    & Z = \hat P_\bp(\bone) = \sum_{j=1}^M \psi^j
\end{align*}

\paragraph{Example.} Consider a set of three Boolean variables, $\{X_1, X_2, X_3\}$. Let their joint probability distribution be given by the state vector
\begin{align*}
    \bp = \begin{Bmatrix}
              0 & 0 & - & \rf \; \psi_1 \\
              - & 1 & 1 & \rf \; \psi_2 \\
              1 & - & 0 & \rf \; \psi_3 \\
              0 & 1 & 0 & \rf \; \psi_4 \\
              1 & 0 & 1 & \rf \; \psi_5
    \end{Bmatrix}
\end{align*}
The probability of the (binary) state vector
\begin{equation*}
    \bs = \begin{Bmatrix}
              1 & - & 0 \\
              - & 0 & 1
    \end{Bmatrix}
\end{equation*}
is given by
\begin{align*}
    & P_\bp(\bs) = \frac{2\,\psi_3 + \psi_1 + \psi_5}{Z}, \\
    & \text{where}\quad Z = 2\,\psi_1 + 2\,\psi_2 + 2\,\psi_3 + \psi_4 + \psi_5
\end{align*}

\paragraph{Defining Probability Distribution.} There are different ways to define a probability distribution. The first approach involves assigning coordinates or exponential factors to each state, as in Eq.~\eqref{eq:prob_mass_coordinatewise}. This is the most explicit way, but it might be impractical.

Alternatively, a probability distribution can be defined as a sum of weighted state vectors using the hybrid notation:
\begin{align*}
    \bp = \bs_1 | \psi_1 + \bs_2 | \psi_2 + \ldots + \bs_m|\psi_m\,,\qquad \bigcup_{i=1}^m \bs_i = \bone
\end{align*}
where the binary vectors $\{\bs_i\}_{i=1}^m$ form a full partition of the state space. In this form, there are only $m$ different exponential factors. These factors are the \emph{degrees of freedom} of our probabilistic model.

Notice that we use \emph{lower indices} to enumerate weighted state vectors. They should not be confused with the upper indices used for coordinate-wise elements.

When using hybrid notation, it is important to take into account the following two aspects:
\begin{itemize}
    \item If not all state vectors $\bs_i$ are mutually orthogonal, there will be states belonging to multiple state vectors. If a state $\statei{j}$ is an element of more than one state vector:
    \begin{equation*}
        \statei{j} \in \bs_{i_1}, \bs_{i_2}, \ldots
    \end{equation*}
    then its exponential factor is calculated as the product of the corresponding factors:
    \begin{equation*}
        \psi^j = \psi_{i_1}\,\psi_{i_2}\, \ldots
    \end{equation*}

    \item The state vectors $\{\bs_i\}$ form a full partition, which means that any state $\bx$ is an element of at least one state vector: $\exists i: \bx \in \bs_i$.
\end{itemize}

\paragraph{Example.} Let
\begin{align*}
    & \bs = \begin{Bmatrix}
                0 & 0 \\
                0 & 1
    \end{Bmatrix} ,\quad
    \bq = \begin{Bmatrix}
              0 & 1 \\
              1 & -
    \end{Bmatrix}
\end{align*}
The distribution defined as
\begin{equation*}
    \bp = \bs | \psi_1 + \bq | \psi_2
\end{equation*}
has the following matrix representation:
\begin{align*}
    & \bp = \begin{Bmatrix}
                0 & 0 & \rf \!\! & \psi_1 \\
                0 & 1 & \rf \!\! & \psi_1 \,\psi_2\\
                1 & 0 & \rf \!\! & \psi_2 \\
                1 & 1 & \rf \!\! & \psi_2
    \end{Bmatrix}
\end{align*}

\paragraph{Logical Hamiltonian.} Up to this point, the system's energy landscape has been described by the vector $\bp$, where each coordinate represents the energy level of a specific state in the configuration space $S_\cE$. To formally describe the local structure of the system, it is convenient to transition to a functional representation.

Let $\cE = \{X_1, \dots, X_N\}$ denote the complete set of variables that parameterises the configuration space $S_\cE$. For any specific global configuration $\bx \in S_\cE$, we define the Hamiltonian function $H(\bx)$ to be the energy level given by the corresponding coordinate of the vector $\bp$:
\begin{align*}
    & H(\statei{i}) = p^i
\end{align*}

Furthermore, for any subset of variables $A \subset \cE$, we adopt the convention that $H(\bx_A)$ indicates a Hamiltonian depending exclusively on the configuration space $S_A$. This Hamiltonian corresponds to the distribution $\bp$ whose causal variables coincide with the subset: $\supp(\bp) = A$. The value of the Hamiltonian $H(\bx_A)$ is the energy level of a state $\bx$ defined by $\bx_A$ on $A$ and extended arbitrarily to the remaining variables in $\cE$.

\subsection{Induced Probability Distribution}
\label{sec:induced-probability-distribution}

In this section we discuss how a distribution can be projected onto a subset of variables.

Let $\bp$ be a probability distribution on the set $\cE$ of $N$ variables. Let $\cK$ be a subset of $\cE$ comprising $m$ variables
\begin{align*}
    & \cK = \{X_{i_1}, \ldots, X_{i_m}\} \subset \cE
\end{align*}
The distribution $\bp$ induces a marginal probability distribution $\bp_\cK$ on the restricted set $\cK$. The induced probability is obtained by summing over all possible configurations of the eliminated variables. This marginalisation involves summation over the exponential factors, rather than the coordinate space. In the space $\cK$, the marginalised columns are removed from the matrix representation of $\bp_\cK$.

The induced probability distribution $\bp_\cK$ on space $S_\cK$ has the same partition function as the original distribution $\bp$ on space $S_\cE$.

\paragraph{Example.} Let $\cE = \{X_1, X_2, X_3\}$, and the probability distribution is given by
\begin{align}
    \label{eq:prob_distribution_example}
    \bp = \begin{Bmatrix}
              0 & 0 & - & \rf \; \psi_1 \\
              - & 1 & 1 & \rf \; \psi_2 \\
              1 & - & 0 & \rf \; \psi_3 \\
              0 & 1 & 0 & \rf \; \psi_4  \\
              1 & 0 & 1 & \rf \; \psi_5
    \end{Bmatrix}
\end{align}
Let $\cK = \{X_2, X_3\}$, which means, we are eliminating variable $X_1$. The induced marginal distribution on $\cK$ becomes
\begin{align*}
    \bp_\cK = \begin{Bmatrix}
                  0 & 0 & \rf \!\! & \psi_1 + \psi_3 \\
                  0 & 1 & \rf \!\! &\psi_1 + \psi_5 \\
                  1 & 0 & \rf \!\! & \psi_3 + \psi_4 \\
                  1 & 1 & \rf \!\! & 2\,\psi_2
    \end{Bmatrix}
\end{align*}
For instance, we observe that the states $\{0\;0\;0\}$ and $\{1\;0\;0\}$ in $\bp$ have exponential factors $\psi_1$ and $\psi_3$, which correspond to their unnormalised probabilities. Summing those probabilities and removing variable $X_1$ yields the state $\{0\;0\}$ with the exponential factor $\psi_1 + \psi_3$, and so on.

The partition functions of $\bp$ and $\bp_\cK$ coincide:
\begin{equation*}
    Z = 2\,\psi_1 + 2\,\psi_2 + 2\,\psi_3 + \psi_4 + \psi_5
\end{equation*}
This is a reflection of the fact that if a probability distribution is normalised, then the marginal distribution is normalised too.

\paragraph{Example: Marginalising over Free Variables.}
Suppose the distribution $\bp$ has some free variables. For example, the following distribution has causal variables $X_1$ and $X_2$, and free variables $X_3$ and $X_4$:
\begin{align*}
    \bp = \begin{Bmatrix}
              0 & 0 & - & - & \rf \; \psi_1 \\
              0 & 1 & - & - & \rf \; \psi_2 \\
              1 & 0 & - & - & \rf \; \psi_3 \\
              1 & 1 & - & - & \rf \; \psi_4
    \end{Bmatrix}
\end{align*}
The induced distribution on the set of causal variables $\cK = \{X_1, X_2\}$ is obtained by summing out the free variables of $\bp$:
\begin{align*}
    \bp_\cK = \begin{Bmatrix}
                  0 & 0 & \rf \; 4\,\psi_1 \\
                  0 & 1 & \rf \; 4\,\psi_2 \\
                  1 & 0 & \rf \; 4\,\psi_3 \\
                  1 & 1 & \rf \; 4\,\psi_4
    \end{Bmatrix}
\end{align*}
It is easy to generalise this observation. In matrix notation, the probability distribution $\bp_\cK$ induced on the set of causal variables of $\bp$ is obtained by a simple transformation: every row of $\bp$ is truncated to $\cK$, and its exponential factor is multiplied by $2^{N - |\cK|}$.

\subsection{Evidence Vector}
\label{sec:evidence-vector}

Consider a set of $N$ variables $\cE = \{X_1, \ldots,X_N\}$. Suppose we are interested in the probability of a specific subset of variables taking certain values. We say that the variables are \emph{observed} or \emph{evidenced}, if their values are fixed. Let's say we observe $k$ variables $\{X_{i_1}, X_{i_2}, \ldots X_{i_k}\}$, that take the values
\begin{equation*}
    X_{i_1} = x_{i_1},\; X_{i_2} = x_{i_2} \ldots,\; X_{i_k}=x_{i_k}
\end{equation*}
where $i_1, \ldots,i_k$ is an arbitrary subset of $1, \ldots, N$. An \textbf{evidence vector} is a state vector that represents the subset of configuration space corresponding to the fixed values of evidenced variables. Below we will use a bold letter $\be$ to designate evidence vector. The evidence vector in matrix representation can be constructed as a single row with numerical entries corresponding to the observed variables, and wildcards otherwise:
\begin{align}
    & \be = \{u_1\ u_2\ \ldots\ u_N\}\,,\quad u_j =
    \begin{cases}
        x_j\,, \quad & \text{if $X_j$ is evidenced} \\
        -\,, & \text{otherwise}
    \end{cases}
\end{align}

For example, in the space of 3 Boolean variables, the evidence vector corresponding to the observation $X_1=1$ and $X_2=0$ becomes:
\begin{equation*}
    \be = \begin{Bmatrix}
              1 & 0 & -
    \end{Bmatrix}
\end{equation*}
Let $\mathcal{K}\subset \cE$ be the set of free variables of the evidence
\begin{equation*}
    \cK = \free(\be)\,,\quad \cK \subseteq \cE
\end{equation*}
and let $S_{\mathcal{K}}$ be the configuration space of~$\mathcal{K}$. We call $S_\cK$ the \textbf{free space} of evidence $\be$.

The probability of $\be$ is calculated the same way as for any other state vector. For instance, for the probability distribution
\begin{align*}
    \bp = \begin{Bmatrix}
              0 & 0 & - & \rf \; \psi_1 \\
              - & 1 & 1 & \rf \; \psi_2 \\
              1 & - & 0 & \rf \; \psi_3 \\
              0 & 1 & 0 & \rf \; \psi_4 \\
              1 & 0 & 1 & \rf \; \psi_5
    \end{Bmatrix}
\end{align*}
and $\be=\{1\;0\;-\}$,  we find
\begin{align*}
    & P_\bp(\be) = \frac{\psi^3 + \psi^5}{Z}
\end{align*}
Notice that this corresponds to the marginalised probability
\begin{align*}
    & P_\bp(\be) = P(X_1=1, X_2=0) = \sum_{X_3\in\{0, 1\}} P(X_1=1, X_2=0, X_3)
\end{align*}

\subsection{Induced Conditional Probability Distribution}
\label{sec:induced-conditional-probability-distribution}

Induced conditional probability combines two projections: onto a subset of variables and onto a subset of the state space.

Let $\bp$ be a probability distribution over~$S_\cE$. Let $\be$ be an evidence vector with the corresponding free space $S_\cK$. We say that there is an induced \textbf{conditional probability distribution}, $\bm{\eta}$, over $S_\cK$, defined as follows. For an arbitrary state $\bx$ of the space $S_\cK$
\begin{equation*}
    \bx \in S_\cK
\end{equation*}
an extended state $\bx' \in S_\cE$ is constructed by appending the fixed values of evidenced variables. Then the coordinate of $\bx$ in $\bet$ coincides with the coordinate of $\bx'$ in $\bp$.

In matrix representation, $\bet$ can be constructed from $\bp$ in the following steps:
\begin{itemize}
    \item Construct $\bet$ as a projection $\bet = \be\,\bp$.
    \item Constrain $\bet$ on $S_\cK$ by:
    \begin{itemize}
        \item Removing rows incompatible with evidence (the ground states of $\be$).
        \item Removing columns corresponding to the evidenced variables (those beyond $\cK$).
    \end{itemize}
\end{itemize}
The resulting vector $\bm{\eta}$ is a well-defined distribution over $S_\cK$, since $\bm{\eta}$ is defined for any truth assignment to variables from $\cK$, and this definition is unique.

For induced conditional probability distribution we will be using the following notation:
\begin{align}
    \bm{\eta} = \be \wedge \bp
\end{align}
We say that $\bet$ is a projection of $\bp$ onto the free space of $\be$.

Notice that, as expected, the partition function of the induced probability distribution differs from the original one. When necessary, we will be using notation indicating which space the function is defined on: $Z^{[\cE]}$ or $Z^{[\cK]}$.

Consider an example. Let $\cE = \{X_1, X_2, X_3\}$, and let the probability distribution be given by the 8-rows vector
\begin{align*}
    & \bp =  \begin{Bmatrix}
                 0 & 0 & 0 & \rf \; \psi^1 \\
                 0 & 0 & 1 & \rf \; \psi^2 \\
                 0 & 1 & 0 & \rf \; \psi^3 \\
                 0 & 1 & 1 & \rf \; \psi^4 \\
                 1 & 0 & 0 & \rf \; \psi^5 \\
                 1 & 0 & 1 & \rf \; \psi^6 \\
                 1 & 1 & 0 & \rf \; \psi^7 \\
                 1 & 1 & 1 & \rf \; \psi^8
    \end{Bmatrix} \\
    & Z^{[\cE]} = \psi^1 + \psi^2 + \psi^3 + \psi^4 + \psi^5 + \psi^6 + \psi^7 + \psi^8
\end{align*}
Let now the evidence be
\begin{equation*}
    \be = \{1\; -\; -\}
\end{equation*}
The induced conditional probability distribution on $\mathcal{K} = \{X_2, X_3\}$ is given by
\begin{align*}
    & \bm{\eta} = \be \wedge \bp =  \begin{Bmatrix}
                                        0 & 0 & \rf \; \psi_5 \\
                                        0 & 1 & \rf \; \psi_6 \\
                                        1 & 0 & \rf \; \psi_7 \\
                                        1 & 1 & \rf \; \psi_8
    \end{Bmatrix} \\
    & Z^{[\cK]} = \psi^5 + \psi^6 + \psi^7 + \psi^8
\end{align*}

An \emph{unnormalised} probability of an evidence vector can be expressed as a partition function of its free space:
\begin{align}
    \hat P_\bp(\be) = \hat P_{\be \wedge \bp}(\bone)
\end{align}
This formula has practical significance, since it provides a recipe for numerical calculation of probabilities. Many computational algorithms in probabilistic logic can be reduced to calculation of probabilities of evidence vectors. For instance, a computation of an arbitrary state vector probability can be expressed in terms of probabilities of evidence vectors (see Sec.~\ref{subsec:state-vector-probability-decomposition}). The conditional probability of an ``output'' variable can also be expressed in terms of the probability of an evidence vector. It is discussed in Section~\ref{sec:conditional-probability}.

More generally, the following identity holds for an arbitrary binary state vector $\bs$:
\begin{align*}
    \hat P_\bp(\be\,\bs) = \hat P_{\be \wedge \bp}(\be \wedge\bs)
\end{align*}
where $\be\wedge\bs$ is the projection of $\bs$ onto the free space of $\be$. If the support of $\bs$ is a subset of the free space of $\be$
\begin{equation*}
    \supp(\bs) \subseteq \cK
\end{equation*}
then this formula simplifies to
\begin{align}
    \label{eq:p_e_s}
    \hat P_\bp(\be\,\bs) = \hat P_{\be \wedge \bp}(\bs)
\end{align}

\subsection{Conditional Probability}\label{sec:conditional-probability}
In practical applications, it is often necessary to calculate the conditional probability of one variable, given full or partial evidence of other variables. Let's consider an example. Suppose we want to find the conditional probability of $X_3$ given that $X_1$ is true:
\begin{equation*}
    P(X_3=1 | X_1=1)
\end{equation*}
To calculate the conditional probability with the state algebra, we first introduce the evidence vector
\begin{equation*}
    \be = \begin{Bmatrix}
              1 & - & -
    \end{Bmatrix}
\end{equation*}
corresponding to the observation that $X_1 = 1$. Next, we introduce two extended evidence vectors, that correspond to the possible values of $X_3$:
\begin{align*}
    &\be_0 = \begin{Bmatrix}
                 1 & - & 0
    \end{Bmatrix} \\
    &\be_1 = \begin{Bmatrix}
                 1 & - & 1
    \end{Bmatrix}
\end{align*}
Using the identity
\begin{align*}
    P(X_3=1 | X_1=1) = \frac{P(X_1=1, X_3=1)}{P(X_1=1, X_3=1) + P(X_1=1, X_3=0)}
\end{align*}
we find
\begin{align*}
    & P(X_3=1 | X_1=1) = \frac{P_\bp(\be_1)}{P_\bp(\be_1) + P_\bp(\be_0)} =
    \frac{\hat P_\bp(\be_1)}{\hat P_\bp(\be_1) + \hat P_\bp(\be_0)}
\end{align*}
Notice that for conditional probability, we only need to calculate unnormalised evidence probability. It does not require the knowledge of the partition function~$Z$ of the entire configuration space.

We can formalise this relationship as follows. Let $\bp$ be the probability distribution on the set of variables $\cE = \{X_1,\ldots,X_N\}$. Without loss of generality, let $X_N$ be the ``output'' variable, whose conditional probability we want to find, given partial or full evidence of the ``input'' variables:
\begin{equation*}
    P(X_N=1 | X_{i_1} = x_{i_1}, \ldots, X_{i_k}=x_{i_k})
\end{equation*}
where $i_1,\ldots,i_k$ is an arbitrary subset of $1,\ldots,N-1$. Let $\be$ be the evidence vector corresponding to the input variables.

Next, we construct two extended evidence vectors, corresponding to the fixed values of the target variable, as follows. Let
\begin{align*}
    \bm{\delta}_0 = \{-\ -\ \ldots\ -\ -\ 0 \}\,,\qquad \bm{\delta}_1 = \{-\ -\ \ldots\ -\ -\ 1 \}
\end{align*}
be the state vectors fixing the target variable only. The extended evidence vectors can be constructed as
\begin{align*}
    &\be_0 = \be\, \bm{\delta}_0\,,\qquad \be_1 = \be\, \bm{\delta}_1
\end{align*}

The conditional probability can now be calculated as
\begin{align}
    \label{eq:cond_prob}
    P(X_N=1 | X_{i_1} = x_{i_1}, \ldots, X_{i_k}=x_{i_k}) = \frac{\hat P_\bp(\be_1)}{\hat P_\bp(\be_1) + \hat P_\bp(\be_0)}
\end{align}

Now, using Eq.~\eqref{eq:p_e_s}, we can rewrite the conditional probability as
\begin{align}
    \label{eq:cond_prob_with_projection}
    &  P(X_N=1 | X_{i_1} = x_{i_1}, \ldots, X_{i_k}=x_{i_k}) =
    \frac{\hat P_{\bet}(\bm{\delta}_1)}{\hat P_{\bet}(\bm{\delta}_1) + \hat P_{\bet}(\bm{\delta}_0)}
\end{align}
where $\bet = \be\wedge\bp$.

\subsection{Probability Additivity and Factorisation}
\label{sec:probability-additivity-and-factorisation}

\paragraph{Probability of a Sum of Orthogonal Vectors}

Let the binary state vector $\bs$ be represented as a sum of \emph{orthogonal} binary vectors
\begin{align*}
    &\bs = \bs_1 + \bs_2\,, \qquad \bs_1 \, \bs_2 = \bzero
\end{align*}

Then the probability of $\bs$ under the probability distribution $\bp$ can be represented as a sum
\begin{align*}
    & \hat P_\bp(\bs_1 + \bs_2) = \hat P_\bp(\bs_1) + \hat P_\bp(\bs_2)
\end{align*}
This is valid for both normalised and unnormalised probabilities.

The intuition behind this formula is straightforward: it represents the probability of the union of the mutually exclusive outcomes (non-overlapping regions of the configuration space).

\paragraph{Sum of Distributions.}
Suppose that the probability distribution $\bp$ is represented as a sum of distributions
\begin{equation*}
    \bp = \sum_{i=1}^m \bp_i
\end{equation*}
Addition in the coordinate space is equivalent to multiplication in the probability space. Indeed, according to the rule of multiplication of exponential factors (Eq.~\eqref{eq:multiplication_psi_factors}), the unnormalised probability of any state $\bx$ can be represented as a product
\begin{align}
    & \hat P_\bp(\bx) = \prod_{i=1}^m \hat P_{\bp_i}(\bx)
\end{align}

\paragraph{Sum of Independent Distributions.} Suppose the probability distribution on a system of $N$ variables is given by the sum of two \emph{independent} state vectors
\begin{align*}
    & \bp = \bp_1 + \bp_2\,, \qquad \bp_1 \wr \bp_2
\end{align*}
Recall that state vectors are independent if they have non-overlapping support:
\begin{equation*}
    \bp_1 \wr \bp_2 \quad \iff \quad \supp(\bp_1) \cap \supp(\bp_2) = \varnothing
\end{equation*}

In this case, the partition function factorises as follows:
\begin{align}
    \label{eq:partition_factorised}
    & \hat P_{\bp_1 + \bp_2}(\bone) = \frac{\hat P_{\bp_1}(\bone) \, \hat P_{\bp_2}(\bone)}{2^N}
\end{align}

\begin{proof}

    Suppose that $\bp$ can be represented as a sum of independent components
    \[
        \bp = \bp_1 + \bp_2\,,\qquad \supp(\bp_1) = \balpha_1\,,\quad \supp(\bp_2) = \balpha_2\,,\quad
        \balpha_1 \cap\balpha_2 = \varnothing
    \]
    The supports may not cover the entire $\cE$:
    \begin{equation*}
        \balpha_1 \cup\balpha_2 \subseteq \cE\,,\quad \alpha_1 + \alpha_2 \leq N\,,\quad\text{where}\quad
        |\balpha_1| = \alpha_1\,,\quad |\balpha_2| = \alpha_2\,,\quad |\cE| = N
    \end{equation*}
    We use the hybrid notation to represent $\bp_1$ and $\bp_2$:
    \begin{align*}
        \bp_1 = \begin{Bmatrix}
                    \bs_{1} | \varphi_{1} \\
                    \bs_{2} | \varphi_{2} \\
                    \vdots \\
                    \bs_{m} | \varphi_{m}
        \end{Bmatrix},\qquad
        \bp_2 = \begin{Bmatrix}
                    \bq_{1} | \psi_{1} \\
                    \bq_{2} | \psi_{2} \\
                    \vdots \\
                    \bq_{n} | \psi_{n}
        \end{Bmatrix},\qquad m = 2^{\alpha_1}, \quad n = 2^{\alpha_2}
    \end{align*}
    In this representation, vectors $\bp_1$ and $\bp_2$ are fully expanded on causal columns, and fully reduced on the free columns. Every vector $\bs_{j}$ contains exactly one row which has $\alpha_1$ numeric entries, and $N - \alpha_1$ wildcards. Consequently, $\bp_1$ has exactly $m=2^{\alpha_1}$ rows in this representation, and every row has $2^{N - \alpha_1}$ states in it. Similarly, $\bp_2$ has $n=2^{\alpha_2}$ rows, each comprising $2^{N - \alpha_2}$ states.

    The partition functions of $\bp_1$ and $\bp_2$ are easy to calculate:
    \begin{align}
        \label{eq:partition_p1}
        & \hat P_{\bp_1}(\bone) = 2^{N-\alpha_1} \sum_{i=1}^m \varphi_{i} \\
        \label{eq:partition_p2}
        & \hat P_{\bp_2}(\bone) = 2^{N-\alpha_2} \sum_{j=1}^n \psi_{j}
    \end{align}
    The multipliers $2^{N-\alpha_1}$ and $2^{N-\alpha_2}$ are due to the presence of wildcards in every row.

    Now let's calculate the partition function of $\bp$. First we observe that the sets $\{\bs_{i}\}$ and $\{\bq_{j}\}$ form the full orthogonal partitioning of the state space:
    \begin{align*}
        \sum_{i=1}^m \bs_{i} = \sum_{j=1}^n \bq_{j} = \bone
    \end{align*}
    Thus, we can write
    \begin{align*}
        & \bp_1 = \sum_i \bs_{i} | \varphi_{i} = \sum_i \bs_{i}\,\bone  | \varphi_{i} =
        \sum_{ij} \bs_{i}\,\bq_{j}  | \varphi_{i} \\
        & \bp_2 = \sum_j \bq_{j} | \psi_{j} = \sum_j \bq_{j}\,\bone  | \psi_{j} =
        \sum_{ij} \bs_{i}\,\bq_{j}  | \psi_{j}
    \end{align*}
    and hence
    \begin{align*}
        &\bp_1 + \bp_2 = \sum_{ij} \bs_{i}\,\bq_{j}  | \varphi_{i}\, \psi_{j}
    \end{align*}
    Since all pairs of rows $\bs_{i}$ and $\bq_{j}$ are independent, they all intersect. Every intersection $\bs_{i}\,\bq_{j}$ can be represented as a single row with $\alpha_1 + \alpha_2$ numerical entries and $N - \alpha_1 - \alpha_2$ wildcards\footnote{
        This intersection corresponds to the product of t-objects. See section~5.5 in~\cite{lesnik2025}.
    }. The exponential factor of this row will be $\varphi_{i}\, \psi_{j}$. Thus, the partition function of $\bp_1 + \bp_2$ can be calculated as
    \begin{align}
        \label{eq:partition_p1_p2}
        & \hat P_{\bp_1 + \bp_2}(\bone) = 2^{N - \alpha_1 - \alpha_2} \sum_{ij} \varphi_{i}\,\psi_{j}
    \end{align}
    Combining Eqs.~\eqref{eq:partition_p1}, \eqref{eq:partition_p2} and \eqref{eq:partition_p1_p2} we obtain Eq.~\eqref{eq:partition_factorised}.
\end{proof}

\paragraph{Non-interacting Hamiltonians.} The factorisation of the partition function can also be understood from the perspective of statistical mechanics.

We consider a partition of the system into non-overlapping subsystems. Specifically, let $A, B \subset \cE$ be disjoint subsets of variables such that $A \cap B = \varnothing$. Assuming these subsystems are non-interacting, the total Hamiltonian decomposes into local, independent components:

\begin{equation*}
    H(\bx) = H_1(\bx_A) + H_2(\bx_B)
\end{equation*}

Because the Hamiltonian splits additively, the corresponding Boltzmann weights factorise. Consequently, the partition function $Z$ can be decomposed into independent sums over the subsystems. Defining $C = E \smallsetminus (A \cup B)$ as the subset of any remaining uncoupled variables, the partition function is given by:

\begin{equation*}
    Z = \sum_{\bx} \exp[-H(\bx)] = \left( \sum_{\bx_A} \exp[-H_1(\bx_A)] \right)
    \left( \sum_{\bx_B} \exp[-H_2(\bx_B)] \right)
    \left( \sum_{\bx_C} 1 \right)
\end{equation*}

The final term, $\sum_{\bx_C} 1$, simply evaluates to $|S_C|$, the total number of possible states for the uncoupled variables in $C$, contributing a constant degeneracy factor to the partition function.

Now we notice that the summation over the spare variables yields the volume of the corresponding configuration subspace
\begin{align*}
    & \sum_{\{\bx_A\}}e^{-H_1} = \frac{1}{2^{N - |A|}}\,\sum_{\{\bx\}}e^{-H_1} \\
    & \sum_{\{\bx_B\}}e^{-H_2} = \frac{1}{2^{N - |B|}}\,\sum_{\{\bx\}}e^{-H_2} \\
    & \sum_{\{\bx_C\}} 1 = 2^{|S_C|} = 2^{N - |A| - |B|}
\end{align*}
Finally, we obtain:
\begin{align*}
    Z = \frac{1}{2^N}\,\sum_{\{\bx\}}e^{-H_1}\,\sum_{\{\bx\}}e^{-H_2}
\end{align*}
which is equivalent to Eq.~\eqref{eq:partition_factorised}.

\paragraph{Example.} Consider the following probability distributions
\begin{align*}
    & \bp_1 = \begin{Bmatrix}
                  0 & 0 & - & - & - & \rf \; \varphi_{1} \\
                  0 & 1 & - & - & - & \rf \; \varphi_{2} \\
                  1 & 0 & - & - & - & \rf \; \varphi_{3} \\
                  1 & 1 & - & - & - & \rf \; \varphi_{4}
    \end{Bmatrix}, \quad
    \bp_2 = \begin{Bmatrix}
                - & - & - & 0 & 0 & \rf \; \psi_{1} \\
                - & - & - & 0 & 1 & \rf \; \psi_{2} \\
                - & - & - & 1 & 0 & \rf \; \psi_{3} \\
                - & - & - & 1 & 1 & \rf \; \psi_{4}
    \end{Bmatrix}\,
\end{align*}
The corresponding partition functions are:
\begin{align*}
    & \hat P_{\bp_1}(\bone) = 2^3 (\varphi_{1} + \varphi_{2} + \varphi_{3} + \varphi_{4}) \\
    & \hat P_{\bp_2}(\bone) = 2^3 (\psi_{1} + \psi_{2} + \psi_{3} + \psi_{4})
\end{align*}

The sum $\bp_1 + \bp_2$ will contain 16 rows corresponding to intersections of all pairs of rows of $\bp_1$ and $\bp_2$:
\begin{align*}
    & \bp_1 + \bp_2 = \begin{Bmatrix}
                          0 & 0 & - & 0 & 0 & \rf \; \varphi_{1} \, \psi_{1} \\
                          0 & 0 & - & 0 & 1 & \rf \; \varphi_{1} \, \psi_{2} \\
                          \vdots \\
                          0 & 1 & - & 0 & 0 & \rf \; \varphi_{2} \, \psi_{1} \\
                          0 & 1 & - & 0 & 1 & \rf \; \varphi_{2} \, \psi_{2} \\
                          \vdots \\
                          1 & 1 & - & 1 & 1 & \rf \; \varphi_{4} \, \psi_{4}
    \end{Bmatrix}, \quad
\end{align*}
In this distribution every row contains 1 wildcard and comprises 2 states. Thus, the partition function becomes
\begin{align*}
    & \hat P_{\bp_1 + \bp_2}(\bone) = 2 \sum_{ij} \varphi_{i} \, \psi_{j} =
    \frac{\hat P_{\bp_1}(\bone)\,\hat P_{\bp_2}(\bone)}{2^5}
\end{align*}
which is in accordance with Eq.~\eqref{eq:partition_factorised}.

Consider a special case of the factorisation formula. Let $\bp_1$ be a uniform distribution $\bp_1 = \bone | \psi_0$
\begin{equation*}
    \bp = \bone | \psi_0 + \bp_2
\end{equation*}
It is easy to see that
\begin{equation*}
    \hat P_{\bp_1}(\bone) = 2^N \,\psi_0
\end{equation*}
and hence,
\begin{align}
    \label{eq:distribution_with_base}
    & \hat P_{\bp_1 + \bp_2}(\bone) = \psi_0 \,\hat P_{\bp_2}(\bone)
\end{align}
which is an expected result, since $\bp_1$ simply multiplies every exponential factor of $\bp_2$ by $\psi_0$. This multiplication doesn't change the normalised probability though.

\section{Probabilistic Logic}\label{sec:probabilistic-logic}

\subsection{Weighted Rules}

A logical \textbf{formula} or a logical \textbf{rule} on configuration space $S_\cE$ is a Boolean function:
\begin{equation*}
    f: S_\cE \to \{0, 1\}
\end{equation*}
The states in $S_\cE$ are partitioned into two disjoint sets: those for which the formula evaluates to true, and those for which it evaluates to false. Let the binary vector $\bs$ represent the set of states satisfying the formula. We designate this as follows:
\begin{equation*}
    f \sim \bs\,,\qquad \forall \bx \in \bs: \ f(\bx) = 1
\end{equation*}
The set of states falsifying the formula is complementary to $\bs$:
\begin{equation*}
    \bq = \bone - \bs\,,\qquad \forall \bx \in \bq: \ f(\bx) = 0
\end{equation*}
For brevity, we refer to states that satisfy $f$ as \emph{allowed} or \emph{valid}, and to those that falsify $f$ as \emph{prohibited} or \emph{invalid}.

\paragraph{Probabilistic Model.} In probabilistic formulation, we assign the logical rule a weighting factor $\psi$ such that
\begin{equation*}
    0 < \psi < 1
\end{equation*}
and define the following distribution associated with the rule:
\begin{align}
    \label{eq:f_distribution}
    \bp = \begin{Bmatrix}
              \bs         & \rf \; \psi^+ \\
              \bone - \bs & \rf \; \psi^-
    \end{Bmatrix} = \bs|\psi^+ + (\bone - \bs)|\psi^-\,,\quad \text{where} \quad
    \psi^+ = \psi,\quad \psi^- = 1 - \psi
\end{align}
We interpret $\psi^+$ as an unnormalised probability of the valid states, and $\psi^-$ as that of the prohibited states.

The model~\eqref{eq:f_distribution} is not the only way to assign a distribution to a rule. See Sec.~\ref{subsec:equivalent-models} for details.

\paragraph{Deterministic Limit.}
To map probabilistic inference back to deterministic logic, we treat the partition function $Z$ and state probabilities $P(\bx)$ as continuous rational functions parameterised by $\psi$. The transition to deterministic logic is defined analytically via the limit $\psi \to 1$. This limit can be approached parametrically by introducing a temperature $T$, defining the distribution coordinates of prohibited states as $p = E/T$, and their corresponding exponential factors as $\psi^- = \exp(-E/T)$.

The asymptotic limit $\psi\to 1$ is then analogous to the zero-temperature limit of Gibbs measures. In this asymptotic limit, the probability mass concentrates entirely on the logical \emph{ground states} (the deterministically valid configurations).

Furthermore, this parameterised limit provides a strict mathematical test for logical consistency. In the limit $\psi\to 1$, the partition function can be expanded as a polynomial $Z(\psi) = c_0 + \mathcal{O}(\epsilon)$, where $c_0$ is the cardinality of the valid state space, and $\epsilon= 1-\psi$. A deterministic logical limit is well-defined if and only if $\lim_{\psi \to 0} Z(\psi) > 0$. If $c_0 = 0$, indicating that all states violate the rule, the limit yields a $0/0$ singularity, rigorously reflecting that the premise is deterministically inconsistent.

\subsection{A System of Weighted Rules}
\label{sec:a-system-of-weighted-rules}

We consider a set of $m$ rules $f_i$ with associated weighting factors $\psi_i$. Let $\bp_i$ be the distribution corresponding to the $i$-th rule. We define the probabilistic model of the system of rules as
\begin{align*}
    & \bp = \bp_1 + \bp_2 + \ldots + \bp_m
\end{align*}
If $\theta_i$ is the exponential factor of a state $\bx$ in the distribution $\bp_i$, then in the total distribution, the exponential factor will be
\begin{equation*}
    \theta = \theta_1\,\theta_2\ldots\theta_m
\end{equation*}
where $\theta_i = \psi_i$ if the $i$-th formula is satisfied, and $1 - \psi_i$ otherwise.

Notice the following:
\begin{itemize}
    \item If $0 < \theta_i < 1$ then
    \begin{equation*}
        0 < \theta < 1
    \end{equation*}
    \item If any $\theta_i$ vanishes, the entire product becomes zero. This means that if a state is prohibited by a single deterministic rule, it is prohibited by the entire system of rules.
\end{itemize}

The transition to a deterministic system of rules is defined similarly. Because the partition function and state probabilities are continuous functions of the weighting factors $\psi_i$, we can evaluate the limit $\psi_i\to 1$ for some or all $i$. This limit is well-defined only if the partition function remains strictly positive, which is possible only if the set of logical formulas is deterministically satisfiable. In this case, the distribution collapses to the zero-temperature limit, where the entire distribution mass concentrates on the ground states.

\paragraph{Equivalent Form with Background Distribution.}
Let the distribution of every formula be represented as~\eqref{eq:f_distribution}. The entire distribution becomes
\begin{align}
    \label{eq:distribution_set_of_rules}
    & \bp  =  \sum_{i=1}^m \left(\bs_i |\psi_i^+ + (\bone -\bs_i)|\psi_i^-  \right)=
    \sum_{i=1}^m \left(\bone |\psi_i^- + \bs_i | \phi_i \right) = \bone |\phi_b + \sum_{i=1}^m \bs_i | \phi_i
\end{align}
where
\begin{align*}
    & \phi_b  = \psi_1^-\,\psi_2^- \ldots \psi_m^- = (1 - \psi_1)(1 - \psi_2)\ldots (1 - \psi_m)\,,\\
    & \phi_i = \frac{\psi_i^+}{\psi_i^-} = \frac{\psi_i}{1 - \psi_i}\,,\qquad i = 1,\ldots,m
\end{align*}
Notice that some $\phi_i$ can be larger than 1, which corresponds to negative coordinates. This doesn't lead to inconsistency, as the total probability will be offset by the multiplier $\phi_b$. In this representation, the components $\bs_i$ ``borrow'' probability from the background distribution term~$\bone|\phi_b$.

Notice that in representation \eqref{eq:distribution_set_of_rules}, the state vectors $\bs_i$ may not be orthogonal:
\begin{equation*}
    \bs_i \,\bs_j \not = \bzero\qquad \text{for some } i \not = j
\end{equation*}

\subsection{Computing Partition Functions}\label{subsec:computing-partition-function}

As demonstrated previously, computing the probabilities of state vectors reduces to calculating the partition function of projected distributions. Therefore, in the following subsections, we focus exclusively on the calculation of
\begin{equation*}
    \hat P_\bp(\bone)
\end{equation*}
for different representations of $\bp$.

\subsubsection{Complete Orthogonal Representation}

Suppose the distribution is represented by a set of mutually-orthogonal state vectors
\begin{align}
    \bp = \sum_{i=1}^m \bs_i | \psi_i\,,\qquad \text{where} \quad
    \bs_i\,\bs_j = \bzero \quad \text{for } i \not = j\,,\quad
    \sum_{i=1}^m \bs_i = \bone
\end{align}
In this case, the partition function becomes:
\begin{align}
    & \hat P_\bp(\bone) = \sum_{i=1}^m \psi_i\,|\bs_i|
\end{align}
where $|\bs_i|$ is the number of states in $\bs_i$. The calculation of $|\bs_i|$ is straightforward in matrix representation (see Sec.~\ref{subsec:state-vector-size}).

\subsubsection{Orthogonal B-Representation}

Suppose the distribution is represented as
\begin{align}
    \bp = \bone|\psi_b + \sum_{i=1}^m \bs_i|\psi_i
\end{align}
where the set of state vectors is orthogonal but might be incomplete (i.e., not fully partitioning the space $S_\cE$):
\begin{equation*}
    \sum_{i=1}^m \bs_i \subseteq \bone\,,\qquad \bs_i\,\bs_j = \bzero \quad \text{for } i \not = j
\end{equation*}
In this representation, the term $\bone|\psi_b$ acts as a base multiplicative offset for the probabilities of all states. We refer to this representation as \emph{orthogonal b-representation}.

To calculate the partition function, let us first complete the set of state vectors by adding a vector $\bs_0$ which contains all states not comprised by any of the state vectors $\{\bs_i\}_1^m$:
\begin{align}
    &\bs_0 = \bone - (\bs_1 + \ldots + \bs_m) \\
    &\bs_{0} + \sum_{i= 1}^{m} \bs_i = \sum_{i= 0}^{m} \bs_i = \bone\,,\qquad \bs_0\,\bs_i = \bzero\quad \text{for } i>0
\end{align}
Now let's assign weight $\psi_0=1$ to the states of $\bs_0$ and add it to the distribution $\bp$. Obviously, adding $\bs_0|1$ to the distribution has no effect on it, since the new term simply multiplies the exponential factors of the states from $\bs_0$ by 1. The modified distribution becomes
\begin{align}
    \bp = \bone|\psi_b + \sum_{i=0}^m \bs_i|\psi_i  = \bone|\psi_b + \bp'
\end{align}
where
\begin{align*}
    & \bp' = \sum_{i=0}^{m} \bs_i|\psi_i \\
    & \bs_0 = \bone - (\bs_1 + \ldots + \bs_m) \\
    & \psi_0 = 1
\end{align*}

Since all state vectors in $\{\bs_i\}_{i=0}^m$ are orthogonal, the number of states in $\bs_0$ is given by
\begin{equation*}
    |\bs_0| = 2^N - (|\bs_1| + |\bs_2| + \ldots + |\bs_m|)
\end{equation*}
Thus, taking into account Eq.~\eqref{eq:distribution_with_base}, the partition function becomes
\begin{align}
    & \hat P_{\bp} = \psi_b\left[2^N + \sum_{i=1}^m (\psi_i - 1)\, |\bs_i|\right]
\end{align}

\subsubsection{Non-orthogonal B-Representation}

Suppose the distribution can be represented as
\begin{align}
    \bp = \bone|\psi_b + \sum_{i=1}^m \bs_i|\psi_i
\end{align}
where the set of state vectors is neither orthogonal nor complete:
\begin{equation*}
    \bigcup_{i=1}^m \bs_i \subseteq \bone\,,\qquad \bs_i\,\bs_j \not = \bzero \quad \text{for some } i \not = j
\end{equation*}
This is the representation of the probability distribution generated by a set of probabilistic rules (see Eq.~\eqref{eq:distribution_set_of_rules}).

To find the partition function of this distribution, we first need to orthogonalise the set of state vectors $\{\bs_i\}$. This can be done iteratively via the following procedure.

As a first step, we find the orthogonal representation for the first pair of state vectors. Consider the expression
\begin{equation*}
    \bp_{(1)} = \bs_1|\psi_1 + \bs_2|\psi_2
\end{equation*}
Each of the vectors $\bs_1$ and $\bs_2$ can be split into two parts: collinear and orthogonal. The collinear part is given by the product $\bs_1\,\bs_2$, while the orthogonal parts are given by $\bs_1 \smallsetminus \bs_2$ and $\bs_2 \smallsetminus \bs_1$ respectively. We can thus rewrite the expression as
\begin{equation*}
    \bp_{(1)} = (\bs_1\smallsetminus\bs_2)|\psi_1 + \bs_1\,\bs_2 | \psi_1\,\psi_2 + (\bs_2 \smallsetminus \bs_1)|\psi_2
\end{equation*}
The three components in this expression are now mutually orthogonal.

In the next step, we find an orthogonal representation for the expression
\begin{equation*}
    \bp_{(2)} = \bp_{(1)} + \bs_3|\psi_3
\end{equation*}
and so on, until the entire expression $\sum_{i=1}^m \bs_i|\psi_i$ is represented as a sum of orthogonal components.

The algorithm for calculating the set difference $\bs\smallsetminus\bq$ is detailed in~\cite{lesnik2025}, Section~5.5.5.

Following this procedure, the distribution is brought to the orthogonal b-representation considered in the previous section.

\section{Notes on Computation and Optimisation}\label{sec:optimisation}

\subsection{State Vector Probability Decomposition}\label{subsec:state-vector-probability-decomposition}

We note an additional, straightforward formula that is highly useful for implementing probabilistic logic. Suppose we want to find the probability of a binary state vector $\bs$ given the distribution $\bp$:
\begin{equation*}
    \hat P_\bp(\bs)
\end{equation*}

In a compact representation, a state vector can be represented as a sum of its rows, where each row acts as an evidence vector:
\begin{equation*}
    \bs = \sum_i \be_i
\end{equation*}
Because $\bs$ is binary, all its rows are mutually orthogonal:
\begin{equation*}
    \be_i\,\be_j = \bzero\,, \qquad \forall i \not = j
\end{equation*}
The probability of $\bs$ can thus be calculated as:
\begin{align*}
    \hat P_\bp(\bs) = \hat P_\bp \left(\sum_i \be_i\right) = \sum_i \hat P_\bp(\be_i) =
    \sum_i \hat P_{\be_i \wedge \bp}(\bone)
\end{align*}

\subsection{Equivalent Models}\label{subsec:equivalent-models}

In this paper, to build a probability model of a set of logical rules, we chose to assign the distribution~\eqref{eq:f_distribution} to every logical formula. There are many equivalent parameterisations of this model.

Let's consider a distribution defined by a single formula:
\begin{align*}
    \bp = \begin{Bmatrix}
              \bs         & \rf \; \psi_1 \\
              \bone - \bs & \rf \; \psi_2
    \end{Bmatrix}
\end{align*}
The normalised probability of a state $\bx$ is
\begin{align*}
    & \frac{\psi_1}{2^k\,\psi_1 + 2^{N-k}\,\psi_2} = \frac{\omega}{2^k\,\omega + 2^{N-k}}\,,
    \qquad \text{if}\quad \bx\in\bs \\
    & \frac{\psi_2}{2^k\,\psi_1 + 2^{N-k}\,\psi_2} = \frac{1}{2^k\,\omega + 2^{N-k}}\,,
    \qquad \text{if}\quad \bx\not \in\bs \\
    & \text{where} \\
    & k = |\bs|\,, \quad  \omega = \frac{\psi_1}{\psi_2}
\end{align*}
We see that the normalised probability of $\bx$ is uniquely defined by the ratio $\omega=\psi_1/\psi_2$. Hence, we can always formulate equivalent models by multiplying the exponential factors $\psi_1$ and $\psi_2$ by a constant (or an arbitrary function of $\psi_1$ and $\psi_2$). It is straightforward to show that this property extends to a system of multiple weighted rules.

In particular, the following alternative model is common in the literature:
\begin{align*}
    \bp = \begin{Bmatrix}
              \bs         & \rf \!\! & \phi \\
              \bone - \bs & \rf \!\! & 1
    \end{Bmatrix}
\end{align*}
This can be obtained from distribution~\eqref{eq:f_distribution} by dividing the exponential factors by $1-\psi$, yielding:
\begin{equation*}
    \phi = \frac{\psi}{1 - \psi}\,,\qquad \phi > 0 \quad \text{if} \quad 0 < \psi < 1
\end{equation*}
In this parameterisation, the transition to deterministic logic occurs in the limit $\phi\to\infty$.

\subsection{State Vector Size}\label{subsec:state-vector-size}

Let $\bs$ be a binary state vector. Let $|\bs|$ be the number of states that $\bs$ comprises. In a compact matrix notation, $\bs$ can be represented as the sum of its rows:
\begin{equation*}
    \bs = \begin{Bmatrix}
              \br_1 \\
              \br_2 \\
              \vdots \\
              \br_n
    \end{Bmatrix} = \br_1 + \br_2 + \ldots + \br_n\,,\qquad
    \br_i \,\br_j = \bzero \text{ for } i \not = j
\end{equation*}
where $\br_i$ is a single-row state vector. Let $w_i$ be the number of wildcards in $\br_i$. Then obviously:
\begin{equation*}
    |\br_i| = 2^{w_i}
\end{equation*}
and hence:
\begin{align}
    |\bs| = \sum_i 2^{w_i}
\end{align}

\subsection{Optimisation}

\subsubsection{Markov Blanket}\label{subsec:markov-blanket}

Suppose we want to find the conditional probability of a target variable given evidence for certain input variables. Let $\be$ be the evidence vector, and let $\cK$ be the free set of $\be$. According to Sec.~\ref{sec:conditional-probability}, to calculate the conditional probability, we must construct extended evidence vectors.

Without loss of generality, let $X_N$ be the output variable. We designate $\bm{\delta}_1$ as an evidence row with a single numerical entry setting the target variable to 1, and wildcards elsewhere. We similarly define $\bm{\delta}_0$:
\begin{align*}
    & \bm{\delta}_1 = \{-\;-\ldots -\; 1\}\,,\quad \supp(\bm{\delta}_1)  = \{X_N\}\,,\\
    & \bm{\delta}_0 = \{-\;-\ldots -\; 0\}\,,\quad \supp(\bm{\delta}_0)  = \{X_N\}
\end{align*}
The extended evidence vectors are then defined as $\be_0 = \be\,\bm{\delta}_0$ and $\be_1 = \be\,\bm{\delta}_1$.

Equation~\eqref{eq:cond_prob_with_projection} allows us to calculate the conditional probability of $X_N$ given the evidence:
\begin{equation*}
    P(X_N=1 | \text{evidence}) = \frac{\hat P_{\bm{\eta}}(\bm{\delta}_1)}{\hat P_{\bm{\eta}}(\bm{\delta}_1) + \hat P_{\bm{\eta}}(\bm{\delta}_0)}
\end{equation*}
where $\bm{\eta} = \be\wedge\bp$ is the conditional distribution obtained by evidencing out the observed variables. The distribution $\bm{\eta}$ is defined over the variables in $\cK$.

Suppose now that $\bm{\eta}$ can be split into two independent components:
\begin{align*}
    & \bm{\eta} = \bm{\eta}_1 + \bm{\eta}_2\,, \qquad  \bm{\eta}_1\wr\bm{\eta}_2 \\
    & \supp(\bm{\eta}_1) = \bm{\alpha_1}\,,\quad \supp(\bm{\eta}_2) = \bm{\alpha_2}\,,\quad
    \bm{\alpha_1}\cap\bm{\alpha_2} = \varnothing
\end{align*}
Let $\bm{\eta}_2$ be the component containing the output variable:
\begin{equation*}
    X_N \in \bm{\alpha}_2
\end{equation*}
For the conditional probability of $\bdelta_0$ we obtain:
\begin{align*}
    & \hat P_{\bm{\eta}}(\bm{\delta}_0) = \hat P_{\bm{\delta}_0\wedge\bm{\eta}}(\bone) =
    \hat P_{\bm{\delta}_0\wedge\bm{\eta}_1 + \bm{\delta}_0\wedge\bm{\eta}_2}(\bone)
\end{align*}
First, we observe that $\bm{\delta}_0\wedge\bm{\eta}_1$ and $\bm{\delta}_0\wedge\bm{\eta}_2$ are also independent, since:
\begin{align*}
    & \supp(\bm{\delta}_0\wedge\bm{\eta}_1) = \bm{\alpha}_1 \smallsetminus \{X_N\} = \bm{\alpha}_1\\
    & \supp(\bm{\delta}_0\wedge\bm{\eta}_2) = \bm{\alpha}_2 \smallsetminus \{X_N\}\\
    & \Rightarrow \\
    & \supp(\bm{\delta}_0\wedge\bm{\eta}_1)\cap \supp(\bm{\delta}_0\wedge\bm{\eta}_2) = \varnothing
\end{align*}
Therefore:
\begin{align*}
    & \hat P_{\bm{\eta}}(\bm{\delta}_0) = \frac{\hat P_{\bm{\delta}_0\wedge\bm{\eta}_1}(\bone)\;
    \hat P_{\bm{\delta}_0\wedge\bm{\eta}_2}(\bone)}{2^{|\cK|}}
\end{align*}
Next, we note that $X_N$ is a free variable in $\bm{\eta}_1$ since $\bm{\alpha}_1 \cap \{X_N\} = \varnothing$. This allows us to conclude that:
\begin{align*}
    & \hat P_{\bm{\delta}_0\wedge\bm{\eta}_1}(\bone) = \frac12 \,\hat P_{\bm{\eta}_1}(\bone)
\end{align*}
Indeed, $\bm{\eta}_1$ can be represented as a set of rows with numeric entries on $\balpha_1$ and wildcards elsewhere. Each row has $|\cK| - |\balpha_1|$ wildcards and comprises $2^{|\cK| - \alpha_1}$ states. Evidence $\bm{\delta}_0$ fixes one of the free variables, reducing the number of wildcards by one, and consequently reducing the number of states in each row by a factor of 2.

Hence:
\begin{align*}
    & \hat P_{\bet}(\bdelta_0) = k\,\hat P_{\bm{\delta}_0\wedge\bm{\eta}_2}(\bone) = k\,\hat P_{\bm{\eta}_2}(\bm{\delta}_0)\\
    & \text{where}\qquad k = \frac{\hat P_{\bm{\eta}_1}(\bone)}{2^{|\cK| + 1}}
\end{align*}
Similarly, for $\hat P_{\bet}(\bdelta_1)$ we get:
\begin{align*}
    & \hat P_{\bet}(\bdelta_1) = k\,\hat P_{\bm{\eta}_2}(\bm{\delta}_1)
\end{align*}
using the same constant $k$. For the final conditional probability, we obtain:
\begin{align*}
    & P(X_N = 1| \text{evidence}) = \frac{\hat P_{\bm{\eta}_2}(\bm{\delta}_1)}
    {\hat P_{\bm{\eta}_2}(\bm{\delta}_1) + \hat P_{\bm{\eta}_2}(\bm{\delta}_0)}
\end{align*}
It turns out that the conditional probability depends solely on the connected component $\bm{\eta}_2$ of the distribution, which encapsulates the target variable.

The set
\begin{equation*}
    \cK \smallsetminus \{X_N\}
\end{equation*}
is known as the \emph{Markov Blanket} of the target variable $X_N$. Effectively, the Markov Blanket is the set of variables that determine the probability of the target variable after the observed variables are evidenced out. Calculating the conditional probability requires summation over all possible configurations of this Markov Blanket. Consequently, identifying the smallest possible blanket can drastically improve computational performance.

\paragraph{Example.}
On the set of 5 variables $\{X_1, X_2, X_3, X_4, X_5\}$, consider the following pair of logical rules with associated weighting factors:
\begin{align*}
    & f_1: X_1 \vee X_2 \Rightarrow X_3\,,\qquad \text{weight} = \psi_1 \\
    & f_2: X_3 \vee X_4 \Rightarrow X_5\,,\qquad \text{weight} = \psi_2
\end{align*}

The overall probability distribution is given by:
\begin{align*}
    & \bp = \bp_1 + \bp_2\,, \\
    & \bp_1 =
    \left\{
        \begin{array}{ccccccc}
            X_1 & X_2 & X_3 & X_4 & X_5 \\
            \hline
            - & - & 1 & - & - & \rf \!\! & \psi_1 \\
            0 & 0 & 0 & - & - & \rf \!\! & \psi_1 \\
            0 & 1 & 0 & - & - & \rf \!\! & 1 - \psi_1 \\
            1 & - & 0 & - & - & \rf \!\! & 1 - \psi_1
        \end{array}
    \right\}, \\
    & \bp_2 =
    \left\{
        \begin{array}{ccccccc}
            X_1 & X_2 & X_3 & X_4 & X_5 \\
            \hline
            - & - & - & - & 1 & \rf \!\! & \psi_2 \\
            - & - & 0 & 0 & 0 & \rf \!\! & \psi_2 \\
            - & - & 0 & 1 & 0 & \rf \!\! & 1 - \psi_2 \\
            - & - & 1 & - & 0 & \rf \!\! & 1 - \psi_2
        \end{array}
    \right\}
\end{align*}
Here, we use header rows to indicate which column corresponds to which variable.

Suppose we want to compute the conditional probability of $X_5=1$ given the evidence $X_3 = 0$:
\begin{equation*}
    P(X_5=1 | X_3=0)
\end{equation*}
Intuitively, we expect that the rule $f_1$ becomes topologically isolated from the target variable $X_5$ once $X_3$ is evidenced; hence, the conditional probability should be independent of $\psi_1$. Let's demonstrate this mathematically.

The evidence vector is:
\begin{equation*}
    \be = \{-\;-\; 0\;-\;-\}
\end{equation*}
By introducing the two output variable evidences:
\begin{align*}
    & \bm{\delta}_0 = \{-\;-\;-\;-\;0\} \\
    & \bm{\delta}_1 = \{-\;-\;-\;-\;1\}
\end{align*}
the conditional probability can be expressed as:
\begin{equation*}
    P(X_5=1 | X_3=0) = \frac{\hat P_{\bet}(\bm{\delta}_1)}{\hat P_{\bet}(\bm{\delta}_1) + \hat P_{\bet}(\bm{\delta}_0)}
\end{equation*}
where $\bet = \be\wedge \bp$. Let's compute $\bet$. First, we multiply $\bp$ by the evidence:
\begin{align*}
    & \be\,\bp = \be \,\bp_1 + \be \,\bp_2 \\
    & \be\,\bp_1 =
    \left\{
        \begin{array}{ccccccc}
            X_1 & X_2 & X_3 & X_4 & X_5 \\
            \hline
            0 & 0 & 0 & - & - & \rf \!\! & \psi_1 \\
            0 & 1 & 0 & - & - & \rf \!\! & 1 - \psi_1 \\
            1 & - & 0 & - & - & \rf \!\! & 1 - \psi_1
        \end{array}
    \right\},\\
    & \be\,\bp_2 =
    \left\{
        \begin{array}{ccccccc}
            X_1 & X_2 & X_3 & X_4 & X_5 \\
            \hline
            - & - & 0 & - & 1 & \rf \!\! & \psi_2 \\
            - & - & 0 & 0 & 0 & \rf \!\! & \psi_2 \\
            - & - & 0 & 1 & 0 & \rf \!\! & 1 - \psi_2
        \end{array}
    \right\},
\end{align*}
The conditional distribution is obtained by removing the variable $X_3$:
\begin{align*}
    & \bet = \be \wedge \bp = \bet_1 + \bet_2 \\
    &\bet_1 =
    \left\{
        \begin{array}{cccccc}
            X_1 & X_2 & X_4 & X_5 \\
            \hline
            0 & 0 & - & - & \rf \!\! & \psi_1 \\
            0 & 1 & - & - & \rf \!\! & 1 - \psi_1 \\
            1 & - & - & - & \rf \!\! & 1 - \psi_1
        \end{array}
    \right\},\\
    & \bet_2 =
    \left\{
        \begin{array}{cccccc}
            X_1 & X_2 & X_4 & X_5 \\
            \hline
            - & - & - & 1 & \rf \!\! & \psi_2 \\
            - & - & 0 & 0 & \rf \!\! & \psi_2 \\
            - & - & 1 & 0 & \rf \!\! & 1 - \psi_2
        \end{array}
    \right\},
\end{align*}
Notice that after evidencing out the observed variable, the conditional distribution splits into the sum of independent distributions $\bet_1 \wr \bet_2$. Indeed:
\begin{equation*}
    \supp(\bet_1) = \{X_1, X_2\}, \quad \supp(\bet_2) = \{X_4, X_5\} \quad \Rightarrow \quad
    \supp(\bet_1)\cap\supp(\bet_2) = \varnothing
\end{equation*}
Consequently, the conditional probability can be re-expressed as:
\begin{equation*}
    P(X_5=1 | X_3=0) = \frac{\hat P_{\bet_1}(\bm{\delta}_1)\,\hat P_{\bet_2}(\bm{\delta}_1)}
    {\hat P_{\bet_1}(\bm{\delta}_1)\,\hat P_{\bet_2}(\bm{\delta}_1) + \hat P_{\bet_1}(\bm{\delta}_0)\,\hat P_{\bet_2}(\bm{\delta}_0)}
\end{equation*}
We now only need to show that $\hat P_{\bet_1}(\bm{\delta}_1) = \hat P_{\bet_1}(\bm{\delta}_0)$, which allows the corresponding factors to cancel out in the equation above.

Because the target variable $X_5$ is among the free variables of $\bet_1$ (meaning its column in the matrix representation of $\bet_1$ contains only wildcards), the distributions $\bm{\delta}_0\wedge\bet_1$ and $\bm{\delta}_1\wedge\bet_1$ coincide:
\begin{align*}
    & \bm{\delta}_0\wedge\bet_1 =\bm{\delta}_1\wedge\bet_1 =
    \left\{
        \begin{array}{cccccc}
            X_1 & X_2 & X_4 \\
            \hline
            0 & 0 & - & \rf \!\! & \psi_1 \\
            0 & 1 & - & \rf \!\! & 1 - \psi_1 \\
            1 & - & - & \rf \!\! & 1 - \psi_1
        \end{array}
    \right\}
\end{align*}
Consequently, the following probabilities are equal:
\begin{align*}
    & \hat P_{\bet_1}(\bm{\delta}_0) = \hat P_{\bm{\delta}_0\wedge\bet_1}(\bone) = \frac12 \hat P_{\bet_1}(\bone) \\
    & \hat P_{\bet_1}(\bm{\delta}_1) = \hat P_{\bm{\delta}_1\wedge\bet_1}(\bone) = \frac12 \hat P_{\bet_1}(\bone)
\end{align*}
Finally, we obtain:
\begin{equation*}
    P(X_5=1 | X_3=0) = \frac{\hat P_{\bet_2}(\bm{\delta}_1)}
    {\hat P_{\bet_2}(\bm{\delta}_1) + \hat P_{\bet_2}(\bm{\delta}_0)}
\end{equation*}
which confirms that we only need to evaluate $\bet_2$ to calculate the conditional probability of $X_5$.

Let us complete the calculation. We find:
\begin{align*}
    & \bm{\delta}_0 \wedge\bet_2 = \left\{
                                       \begin{array}{cccccc}
                                           X_1 & X_2 & X_4 \\
                                           \hline
                                           - & - & 0 & \rf \!\! & \psi_2 \\
                                           - & - & 1 & \rf \!\! & 1 - \psi_2
                                       \end{array}
    \right\} , \\
    & \bm{\delta}_1 \wedge\bet_2 = \left\{
                                       \begin{array}{cccccc}
                                           X_1 & X_2 & X_4 \\
                                           \hline
                                           - & - & - & \rf \; \psi_2
                                       \end{array}
    \right\}
\end{align*}
Hence:
\begin{align*}
    & \hat P_{\bet_2}(\bm{\delta}_0) = \hat P_{\bm{\delta}_0\wedge \bet_2}(\bone) = 4\,\psi_2 + 4\,(1 - \psi_2)\,, \\
    & \hat P_{\bet_2}(\bm{\delta}_1) = \hat P_{\bm{\delta}_1\wedge \bet_2}(\bone) = 8\,\psi_2
\end{align*}
And we finally arrive at:
\begin{align*}
    & P(X_5=1 | X_3=0) = \frac{8\,\psi_2}{8\,\psi_2 + 4\,\psi_2 + 4(1 - \psi_2)} =
    \frac{2\,\psi_2}{1 + 2\,\psi_2}
\end{align*}

\subsubsection{Distribution Factorisation}

As we have seen, many algorithms in probabilistic inference reduce to computing a partition function over the free space of the evidence. In the worst-case scenario, the complexity of these algorithms scales exponentially with the number of free variables. Although the compact representation of state vectors can drastically reduce practical complexity, it is always possible to construct pathological cases where compactification is inefficient. Factorisation of the distribution can circumvent this, significantly improving performance. While it does not change the asymptotic exponential bound itself, it can dramatically expand the boundary of computationally tractable scenarios.

Consider a case where we need to compute a partition function over a set of $K$ variables, and assume the algorithm's complexity scales exponentially as:
\begin{equation*}
    \mathcal{O}(2^{a\,K})
\end{equation*}
for some constant $a$. Now suppose we find a way to factorise the distribution. For simplicity, assume the factorisation perfectly halves the variables, splitting the distribution into two independent distributions, each with a support of $K/2$ variables. Because the probability factorises, the total computation time is merely twice the time required to evaluate each individual sub-distribution. We thus estimate the new complexity as:
\begin{equation*}
    \mathcal{O}\left(2^{\frac{a}{2}\,K + 1}\right)
\end{equation*}
More generally, splitting the problem into $n$ equal-sized independent distributions yields an asymptotic complexity of:
\begin{equation*}
    \mathcal{O}\left(n\,2^{\frac{a}{n}K}\right)
\end{equation*}

Graph-theoretic methods simplify the identification of these independent distribution components. For a distribution represented as:
\begin{equation*}
    \bp = \bs_1|\psi_1 + \bs_2|\psi_2 + \ldots
\end{equation*}
we can construct an interaction graph where vertices correspond to variables, and edges are drawn such that the support of each state vector $\bs_i$ forms a clique. Within this graph, the independent parts of the probability distribution map directly to disjoint, connected components.

\subsubsection{Vertex Separator}

In this section, we explore how identifying a \emph{vertex separator} can facilitate probability factorisation, making computational algorithms more efficient.

Consider a distribution $\bp$ represented as:
\begin{equation*}
    \bp = \bs_1|\psi_1 + \bs_2|\psi_2 + \ldots
\end{equation*}
As before, we construct a constraint graph where variables are vertices, and each state vector $\bs_i$ forms a clique over $\supp(\bs_i)$.

If the entire graph forms a single connected component, the global probability does not natively factorise. However, it is sometimes possible to identify a small subset of nodes that, if removed, would sever the graph into two or more disconnected components. In graph theory, such a subset is called a \emph{vertex separator}.

Because these separator variables are unobserved, their ``removal'' cannot be achieved by evidencing them out (i.e., we cannot arbitrarily fix their values). Instead, we perform an explicit summation over all possible value assignments for the vertex separator.

Returning to the example from the Markov Blanket section, suppose now that variable $X_3$ is \emph{not} observed. Given the two weighted logical rules:
\begin{align*}
    & f_1: X_1 \vee X_2 \Rightarrow X_3\,,\qquad \text{weight} = \psi_1 \\
    & f_2: X_3 \vee X_4 \Rightarrow X_5\,,\qquad \text{weight} = \psi_2
\end{align*}
we need to compute the marginal probability of the target variable $X_5$:
\begin{equation*}
    P(X_5=1) = \frac{\hat P(X_5 = 1)}{\hat P(X_5 = 1) + \hat P(X_5 = 0)}
\end{equation*}

Because $X_3$ is not evidenced, the rule $f_1$ is not topologically isolated from $X_5$, and the Markov Blanket of $X_5$ encompasses all the preceding variables $\{X_1, X_2, X_3, X_4\}$.

Let's perform an explicit summation over the possible states of variable $X_3$:
\begin{align*}
    & \hat P(X_5=1) = \hat P(X_3 = 0, X_5=1 ) + \hat P(X_3 = 1, X_5=1) \\
    & \hat P(X_5=0) = \hat P(X_3 = 0, X_5=0) + \hat P(X_3 = 1, X_5=0)
\end{align*}
By designating the explicit evidence states for $X_3$ and $X_5$ as:
\begin{align*}
    & \be_0 = \{-\;-\;0\;-\;-\}\,, \\
    & \be_1 = \{-\;-\;1\;-\;-\}\,, \\
    & \bdelta_0 = \{-\;-\;-\;-\;0\}\,, \\
    & \bdelta_1 = \{-\;-\;-\;-\;1\}
\end{align*}
we can rewrite the target probabilities as:
\begin{align*}
    & \hat P(X_5=1) = \hat P_{\be_0\wedge\bp}(\bdelta_1) + \hat P_{\be_1\wedge\bp}(\bdelta_1)\,, \\
    & \hat P(X_5=0) = \hat P_{\be_0\wedge\bp}(\bdelta_0) + \hat P_{\be_1\wedge\bp}(\bdelta_0)
\end{align*}
While this explicit summation over $X_3$ doubles the total number of terms to compute, each resulting conditional distribution ($\be_0\wedge\bp$ and $\be_1\wedge\bp$) now has $X_3$ effectively evidenced out. This allows the distributions to fracture into independent components, a factorisation that can yield a net reduction in overall computational complexity.

\begin{figure}[htbp]
    \centering
    \vspace{1.5em}

    \begin{tikzpicture}[
        standard/.style={circle, draw=black!80, fill=blue!10, very thick, minimum size=11mm, font=\small},
        separator/.style={circle, draw=red!80, fill=red!20, very thick, minimum size=11mm, font=\small}
    ]
        \node[separator] (X3) at (0, 0) {$X_3$};

        \node[standard] (X1) at (-2.5, 1.5) {$X_1$};
        \node[standard] (X2) at (-2.5, -1.5) {$X_2$};

        \node[standard] (X4) at (2.5, 1.5) {$X_4$};
        \node[standard] (X5) at (2.5, -1.5) {$X_5$};

        \draw[thick] (X1) -- (X2);
        \draw[thick] (X1) -- (X3);
        \draw[thick] (X2) -- (X3);

        \draw[thick] (X3) -- (X4);
        \draw[thick] (X3) -- (X5);
        \draw[thick] (X4) -- (X5);

    \end{tikzpicture}
    \vspace{1.5em}
    \caption{Vertex separator $X_3$ bridging two components.}\label{fig:graph_separator}
\end{figure}
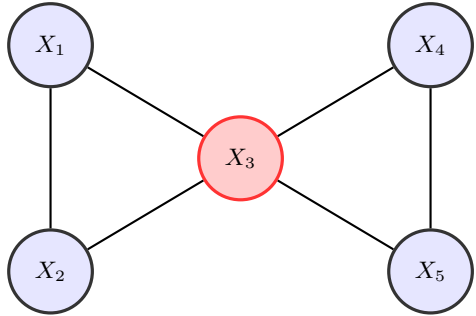

Generally, identifying a vertex separator that contains $m$ variables requires performing $2^m$ explicit summations. If the identified separator is too large, the structural gain from splitting the distribution is outweighed by the exponentially increasing number of terms.

Let us formalise the net performance gain of this algorithm. Suppose the original, un-factorised distribution spans $K$ variables with a worst-case computational complexity of $\mathcal{O}(2^{aK})$, where the parameter $a$ is dictated by the efficiency of the underlying state algebraic transformations.

Suppose a vertex separator of size $m$ splits the remaining variables into disjoint, independent components, and let $K_{\max}$ denote the number of variables in the largest resulting component. By explicitly summing over the separator, the total complexity becomes strictly bounded by the $2^m$ evaluations of this largest component:
\begin{equation*}
    \mathcal{O}(2^m \cdot 2^{a K_{\max}}) = \mathcal{O}(2^{m + a K_{\max}})
\end{equation*}
This factorisation yields a theoretical performance improvement if:
\begin{equation*}
    m < a(K - K_{\max})
\end{equation*}
Note that if the separator achieves a near-perfect bisection ($K_{\max} \approx K/2$), this threshold condition simplifies to $m < \frac{aK}{2}$.

In practical implementations, the computational overhead of locating the separator must also be accounted for. Finding an optimal, strictly minimum vertex separator is known to be NP-hard; thus, polynomial-time heuristic algorithms running in $\mathcal{O}(P(K))$ are typically deployed. Consequently, the true algorithmic complexity is $\mathcal{O}(P(K) + 2^{m + a K_{\max}})$. Furthermore, by recursively applying this vertex separator strategy to the resulting sub-components, one essentially constructs a tree decomposition of the constraint graph. In this fully factorised limit, the complexity of exact inference in State Algebra scales with the treewidth of the structural graph.

Because finding a strong vertex separator heuristically can introduce latency, software implementations often employ an ``algorithm portfolio'' approach. By executing the separator search in parallel with a standard brute-force probability calculation, the system can dynamically terminate the brute-force thread and seamlessly switch to exact inference using the faster, factorised components if a viable separator is discovered first.

\section{Probabilistic Rule Models}\label{sec:probabilistic-rule-models}

The theoretical framework for a system of weighted rules introduced in Section~\ref{sec:a-system-of-weighted-rules} provides a rigorous basis for constructing \textit{Probabilistic Rule Models} (PRMs). In a PRM, the global probability distribution is composed of a collection of local, modular rules that collectively define the energy landscape of the configuration space. This architecture is particularly well-suited for building decisioning systems that require both high predictive accuracy and human-verifiable logic.

In practical applications, these models often take the form of an implication-based structure where a set of input features $\mathcal{X} = \{X_1, X_2, \dots, X_k\}$ implies an output variable $Y$. For a deterministic rule $f: \left(\bigwedge X_i\right) \Rightarrow Y$, the probabilistic version assigns a weight $\psi$ representing the confidence or reliability of that specific implication. This setup allows the model to handle the noise and contradictions inherent in real-world data -- a task that traditional deterministic \textit{Business Rule Engines} (BREs) struggle to perform without manual and often exhaustive consistency checks.

A primary advantage of this approach is the ease with which logical rules can be extracted from empirical data. When the causal structure of a domain is unknown, \textit{association rule mining} techniques can be employed to discover significant implications. Algorithms such as \textit{Apriori}~\cite{agrawal1994fast} and \textit{FP-Growth}~\cite{han2000mining} are routinely used to identify frequent itemsets and generate rules based on metrics like support and confidence. These mined rules are then mapped to the probabilistic state algebra by using the derived confidence levels as an initial basis for the weighting factors $\psi_i$. To ensure the model's predicted probabilities precisely align with the actual probabilities observed in the data, these weights can be further calibrated using standard gradient descent methods. This automated discovery and calibration process can be seamlessly refined by incorporating expert knowledge, empowering human specialists to manually adjust weights or introduce high-level domain constraints that purely data-driven processes often miss.

The utility of PRMs has already been demonstrated in several high-stakes domains. In the context of healthcare, PRMs were successfully utilised to predict the onset of viral symptoms using data generated from wearable technology~\cite{dhaese-et-al:2021}. In that study, the intrinsic modularity of the rules allowed disparate physiological signals to be integrated into a cohesive, highly interpretable predictive model. Similarly, within the financial sector, PRMs have been deployed as \textit{diagnostic} or \textit{correction layers} to interpret and adjust for structural concept drift in post-crisis environments~\cite{lesnik2025probabilistic}. In such cases, the PRM functions as a transparent analytical overlay on top of an existing black-box model, providing human-readable explanations for model adjustments while retaining strict underlying probabilistic rigour.

By framing complex decisioning as a summation of weighted logical components within a vector space, PRMs bypass the traditional historical trade-off between ``black-box'' performance and ``white-box'' explainability. The resulting system is not only inherently robust to logical inconsistencies, but it also remains low-dimensional and structurally accessible. Rules can be added, removed, or independently updated without forcing a complete, expensive retraining of the global probability distribution.

\section{Conclusion and Outlook}\label{sec:conclusion}

In this paper, we have introduced a \textit{Probabilistic State Algebra} that formalises logical inference as a sequence of discrete, matrix-based operations. By interpreting the underlying configuration space through the lens of coordinate-wise Hadamard products and energy potentials, we have demonstrated that advanced probabilistic reasoning can be successfully decoupled from the structural dependencies of traditional graph-based compilation. This unique approach effectively bypasses the rigid variable ordering constraints characteristic of canonical decision diagrams, offering a highly parallelisable and scalable alternative for exact inference within \textit{Markov Random Fields}.

A significant downstream contribution of this mathematical work is the formal development of \textit{Probabilistic Rule Models} (PRMs), which provide a uniquely modular and interpretable framework for automated decisioning. Unlike standard deterministic rule engines, PRMs natively absorb logical inconsistencies and overlapping constraints via probabilistic weighting, profoundly simplifying the maintenance of complex industrial systems. By enabling the seamless mathematical integration of data-mined implications alongside expert-driven constraints, the algebra provides native support for robust human-in-the-loop workflows. This ensures that automated systems can remain auditable, transparent, and defensible in high-stakes environments such as global finance and healthcare.

Looking ahead, several promising avenues for extension remain. While the current work deliberately focuses on propositional logic, the fundamental algebraic structures presented here can theoretically be extended to higher-order logic paradigms. Mapping quantifiers and predicates directly into the state-vector representation would allow researchers to model significantly more complex relational dependencies, all while remaining within the exact same linear-algebraic framework. Furthermore, the transition towards infinite systems -- where the number of variables or the overall domain of truth assignments becomes uncountable -- presents a compelling and challenging theoretical frontier. Preliminary steps in this specific direction were previously established in the deterministic context~\cite{lesnik2025}, suggesting that the algebra can ultimately be generalised to handle continuous state spaces or infinite-dimensional configuration spaces by employing advanced functional analytic methods.

Finally, the highly computational nature of the algebra suggests immediate and lucrative opportunities for modern hardware acceleration. The ``embarrassingly parallel'' nature of the Hadamard product, combined seamlessly with the ultra-sparse representations afforded by \textit{t-objects}, makes this specific computational framework absolutely ideal for direct deployment on Tensor Processing Units (TPUs) and specialised deep-learning AI accelerators. Future engineering research will explore the deep integration of these algebraic primitives directly into large-scale deep learning architectures, potentially serving as a rigorous symbolic ``reasoning layer'' that bridges the formidable flexibility of modern neural networks with the rigorous, explainable interpretability of pure probabilistic logic.

\bibliography{bibliography}

\end{document}